\documentclass[10pt,journal,compsoc]{IEEEtran}
\ifCLASSOPTIONcompsoc
  \usepackage[nocompress]{cite}
\else
  \usepackage{cite}
\fi
\ifCLASSINFOpdf
   \usepackage[pdftex]{graphicx}
\else
   \usepackage[dvips]{graphicx}
\fi
\usepackage{algorithm}
\usepackage{algorithmic}

\usepackage{array}
\usepackage{url}

\usepackage{multirow}
\usepackage{amsmath,amssymb,amsfonts}

\begin{document}
%
\title{Deep-AIR: A Hybrid CNN-LSTM Framework for Air Quality Modeling in Metropolitan Cities}
%
%
%
%

\author{Yang~Han\IEEEauthorrefmark{2},
        Qi~Zhang\IEEEauthorrefmark{2},
        Victor~O.K.~Li\IEEEauthorrefmark{1},~\IEEEmembership{Life~Fellow,~IEEE},
        and~Jacqueline~C.K.~Lam\IEEEauthorrefmark{1},~\IEEEmembership{Member,~IEEE}
\IEEEcompsocitemizethanks{\IEEEcompsocthanksitem The authors are with the Department of Electrical and Electronic Engineering, The University of Hong Kong, Pok Fu Lam, Hong Kong.\protect\\
E-mail: \{yhan, zhangqi, vli, jcklam\}@eee.hku.hk
}
\thanks{
\IEEEauthorrefmark{2} Authors with equal contributions.
}
\thanks{
\IEEEauthorrefmark{1} Corresponding authors.%
}}

\IEEEtitleabstractindextext{%
\begin{abstract}
Air pollution has long been a serious environmental health challenge, especially in metropolitan cities, where air pollutant concentrations are exacerbated by the street canyon effect and high building density. Whilst accurately monitoring and forecasting air pollution are highly crucial, existing data-driven models fail to fully address the complex interaction between air pollution and urban dynamics. Our Deep-AIR, a novel hybrid deep learning framework that combines a convolutional neural network with a long short-term memory network, aims to address this gap to provide fine-grained city-wide air pollution estimation and station-wide forecast. Our proposed framework creates 1x1 convolution layers to strengthen the learning of cross-feature spatial interaction between air pollution and important urban dynamic features, particularly road density, building density/height, and street canyon effect. Using Hong Kong and Beijing as case studies, Deep-AIR achieves a higher accuracy than our baseline models. Our model attains an accuracy of 67.6\%, 77.2\%, and 66.1\% in fine-grained hourly estimation, 1-hr, and 24-hr air pollution forecast for Hong Kong, and an accuracy of 65.0\%, 75.3\%, and 63.5\% for Beijing. Our saliency analysis has revealed that for Hong Kong, street canyon and road density are the best estimators for NO\textsubscript{2}, while meteorology is the best estimator for PM\textsubscript{2.5}.
\end{abstract}

}

\maketitle

\IEEEdisplaynontitleabstractindextext

%
\IEEEpeerreviewmaketitle

\IEEEraisesectionheading{\section{Introduction}\label{sec:introduction}}

%
%
%
%
\IEEEPARstart{R}{apid} socio-economic development and urbanization have led to severe air quality deterioration in many parts of the world over the past decades, especially in developing countries such as China and India. Many adverse health outcomes, such as respiratory and cardiovascular diseases \cite{guan2016impact}, mental health issues \cite{xue2019declines}, and more recently, Covid-19 infection and mortality \cite{copat2020role}, have been associated with the rise of air pollution levels. Providing city-wide air quality information has significant implications for promoting healthy urban living and improving the well-being of citizens. On the one hand, accurate air pollution information can inform citizens (especially children and the elderly) in different parts of the city so as to reduce the health risks of air pollution exposure and improve their quality of life. On the other hand, fine-grained air quality estimates can facilitate evidence-based environmental and public health policymaking, such as setting out traffic control plans in highly polluted areas. However, air quality monitoring stations are often geographically sparse in a city (e.g., only 18 monitoring stations in Hong Kong, covering an area of more than one thousand square kilometers with more than 7.5 million residents), making it extremely challenging to provide accurate and timely air pollution reporting covering every part of the city.

Researchers have proposed many methods for air quality modeling within a city \cite{jerrett2005review}. These urban air quality models have focused on two approaches, including the physical-based and data-driven approaches. Physical-based models utilize numerical methods to describe the air pollution process \cite{lateb2016use}, whereas data-driven models exploit patterns learned from historical air pollution data through statistics and machine learning \cite{rybarczyk2018machine}. More recently, big data and deep learning techniques have pushed the frontiers of the traditional data-driven approach and have achieved state-of-the-art performance in urban air quality modeling \cite{cheng2018neural,yi2020predicting,chen2019deep}. Moreover, urban air quality models have centered on two objectives. The first objective is estimating air pollution in areas without monitoring stations (also referred to as fine-grained air pollution estimation) \cite{cheng2018neural}. The second objective is predicting air pollution in the future (also referred to as air pollution forecast) \cite{yi2020predicting}. Although many studies have investigated the problem of estimating city-wide fine-grained air pollution in the current hour or forecasting air pollution levels for monitoring stations in the next hours, few data-driven models have attempted to achieve these two objectives jointly \cite{chen2019deep, zhao2017incorporating}.

Until now, it remains a challenge to provide fine-grained air quality estimations throughout the city and air pollution forecasts at monitoring stations using deep learning. Compared to other research areas where deep learning algorithms have been successfully adopted, deep learning-based urban air quality modeling is limited by the incompleteness of historical data and the geographical sparsity of air pollution monitoring stations. There are many missing values in the dataset in both the temporal and spatial dimensions. The resultant lack of training data and the noise brought by missing values severely harm the performance of air quality modeling. A large amount of readily available urban proxy data (also referred to as urban dynamics) can be utilized by the deep learning models to address the missing/sparse data issue. These urban dynamics and their interactions can directly or indirectly influence the spatio-temporal variation of air pollution levels within cities.

\subsection{Factors Affecting Air Quality in Urban Environments}
Previous deep learning studies have identified a number of factors that can affect air pollution levels in urban environments \cite{yi2020predicting,chen2019deep,lin2018exploiting}. These factors can generally be categorized as follows: (1) factors related to the reaction, diffusion, or transport of air pollutants, such as meteorology (and weather forecast) \cite{yi2020predicting} and urban morphology, e.g., points of interests (POIs) such as buildings and parks \cite{lin2018exploiting}, (2) direct emission sources represented by factors such as traffic conditions, road networks, factory emissions, and POIs such as factories \cite{chen2019deep}, (3) secondary sources due to chemical reactions between multiple pollutants such as NO\textsubscript{2} and O\textsubscript{3} \cite{yi2020predicting}, and (4) fixed effects representing the unobserved factors (such as human activities) that contribute to the seasonal variation of air pollution such as day of the week \cite{yi2020predicting}.

Air pollution and other urban dynamics data are often correlated to each other, and they tend to interact in a complicated way across different urban environments. For instance, the severity of PM\textsubscript{2.5} pollution in Beijing, China, was highly influenced by meteorological conditions \cite{liang2016pm2}. A study carried out at the city of Madrid, Spain, showed that meteorological factors, including wind speed and cloud type, have strong influences on CO, NO, NO\textsubscript{2}, and O\textsubscript{3} concentrations, whereas local traffic conditions have a minimal impact on PM\textsubscript{10} concentration \cite{lana2016role}. Another study revealed that in Suzhou, China, motor vehicle emission is the most influential factor contributing to NO\textsubscript{2} levels \cite{costabile2006preliminary}. In addition to meteorology and traffic conditions, the street canyon effect, a consequence of the complex interaction between air pollution, meteorology, traffic conditions, and urban morphology (road networks and building geometries), can often be observed in urban environments. A street canyon was initially defined as a relatively narrow street with buildings lined up continuously along both sides, and now the term is also used to represent larger urban streets that are not necessarily flanked by buildings continuously on both sides \cite{vardoulakis2003modelling}. Previous studies have identified high air pollution levels inside street canyons in high-density urban areas in Hong Kong \cite{shi2016developing} and Beijing \cite{fu2017effects}.


\begin{figure}[!t]
    \centering
    \includegraphics[width=0.45\textwidth]{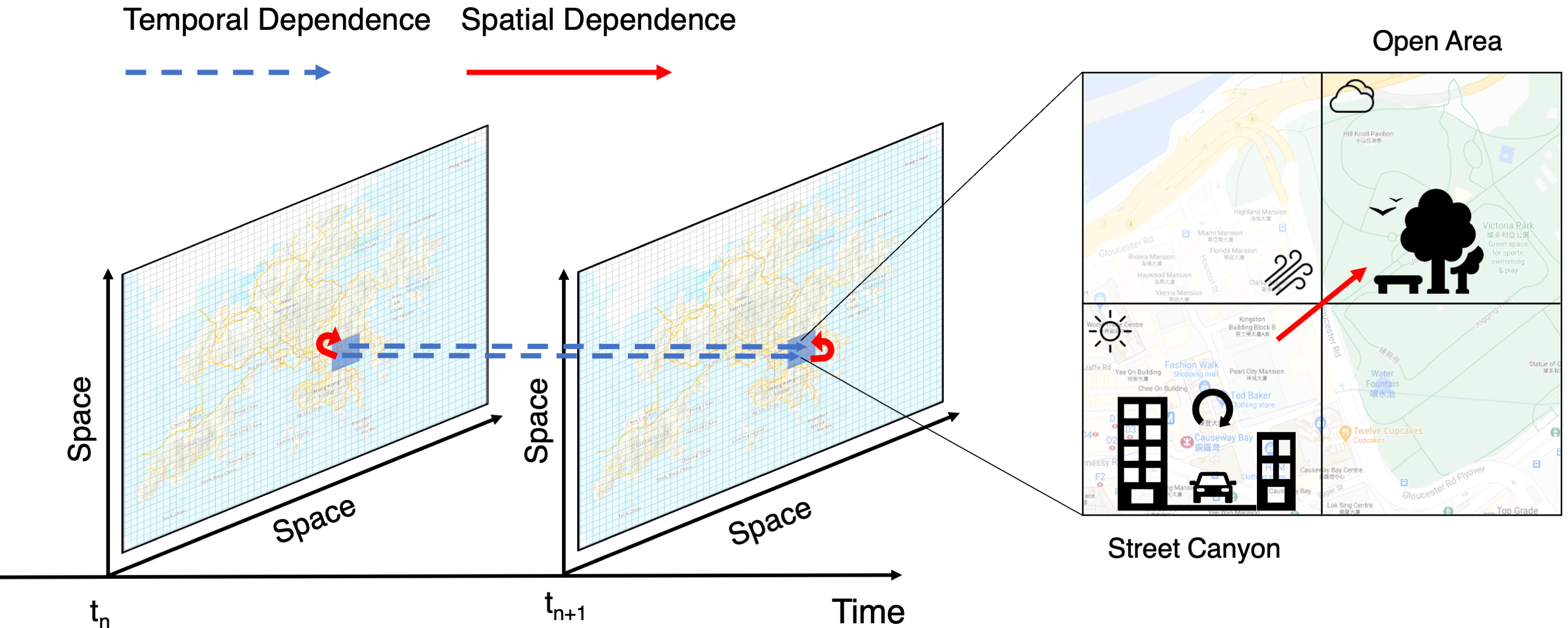}
    \caption{An Illustration for the Temporal and Spatial Correlation of Urban Dynamics Data}
    \label{fig:illstration}
\end{figure}

\subsection{Motivation and Research Significance}
Air pollution and other urban dynamics data are both temporally and spatially correlated. The complex temporal and spatial interaction between air pollution and other urban dynamics must be addressed to capture the variation of air pollution levels across different urban environments at a fine-grained level. Figure \ref{fig:illstration} illustrates the temporal and spatial correlation of air pollution and other urban dynamics data in two urban environments (a street canyon and an open area). First, the air pollution levels at one location can be correlated with the historical air pollution observations in the same location, depending on the local conditions such as traffic emissions and meteorological conditions. For example, we can observe a significant increasing trend in air pollution levels in a street canyon during peak hours due to the complex interaction between traffic emissions, meteorological conditions (such as temperature and solar radiation), and photochemical reactions. Second, the air pollution levels at one location are often dependent on the surrounding area due to the diffusion and transport of air pollutants. For example, although the air quality levels in an open area with no vehicle emissions tend to be better compared to a street canyon, we can observe a rapid deterioration in air quality in the open area due to the transport of air pollutants originated from nearby areas, depending on factors such as wind speed and direction.

However, few deep learning studies have investigated the complex spatial interaction between air pollution and important urban dynamics in high-density urban areas such as traffic conditions and street canyons (see Section \ref{sec:related_work} for a detailed review of related work). Earlier deep learning-based research considered the temporal correlation of air quality and urban dynamics using sequential models, such as recurrent neural networks (RNNs) and their variants long short-term memory (LSTM) networks \cite{li2017long}. More advanced models addressed the spatial dependence between air pollution and other urban dynamics, utilizing spatial models such as convolutional neural networks (CNNs) \cite{wen2019novel, zhang2020deep} and graph convolutional networks (GCNs) \cite{lin2018exploiting, qi2019hybrid}. However, these urban air quality models did not incorporate important features indicative of the street canyon effect in urban areas, such as building density, building height, and street canyon. Until now, given that these models are yet to address the street canyon effect explicitly, it remains unknown which advanced spatial model structures can better capture the characteristics of air pollution and urban dynamics data, and whether or not the street canyon-related features are important to the spatio-temporal prediction of air pollution in urban areas using deep learning.

This study aims to fill this gap by developing Deep-AIR, a hybrid deep learning framework for providing fine-grained air pollution estimations at the city-wide level in the current hour and air pollution forecasts at monitoring stations in the next hours. Our proposed framework, taking city-wide urban dynamics as image-like data, incorporates a CNN component with 1x1 convolution layers to extract the spatial feature representation and an RNN component implemented by the LSTM model to learn the temporal correlation of the extracted features. The 1x1 convolution layers are adopted to strengthen the learning of cross-feature spatial representation between air pollution and various important urban dynamic features, including meteorology, traffic, and urban morphology (particularly road density, building density/height, and street canyon features). This study evaluates the results of our proposed deep learning framework for urban air quality modeling, using data collected in Beijing and Hong Kong. The main contributions of our work are listed as follows.

\begin{itemize}
\item To the best of our knowledge, this is the first deep learning model that includes a CNN component in learning the spatial variation of air pollution data, while utilizing 1x1 convolution layers to capture the spatial interaction between air pollution and various urban dynamic features, including meteorology, traffic, and urban morphology (particularly road density, building density/height, and street canyon features).
\item With the incorporation of domain-specific features related to the street canyon effect, our proposed hybrid CNN-LSTM model has achieved the best performance compared to the baseline models, including statistical and deep learning models.
\item We provide a saliency analysis of input features to improve the interpretability of the proposed framework and reveal the importance of domain-specific features in urban air quality modeling, including features related to traffic conditions and street canyons.
\end{itemize}

Although this study uses Hong Kong and Beijing as the case studies, our proposed urban air quality prediction framework can be transferred to other highly populated and polluted areas/countries where urban proxy data are readily available, such as India. Our novel methodology, capturing the complex interaction of domain-specific spatio-temporal features, can potentially contribute to a wide range of interdisciplinary research topics in urban computing and social sciences, such as city-wide crowd/traffic flows prediction and fine-grained wealth estimation.

Before this study, we published our earlier results in a preprint article \cite{zhang2020deep}. This study extends our previous work by (1) providing a comprehensive literature review to include the most recent studies, (2) utilizing more domain-specific features, including road networks and building geometries, to address the street canyon effect, and (3) providing a saliency analysis to reveal the most important domain-specific features in predicting urban air quality. The rest of this paper is organized as follows. Section \ref{sec:related_work} reviews related work in urban air quality modeling and highlights the added value of this study. Section \ref{sec:method} illustrates our proposed novel methodology for fine-grained air pollution estimation at the city-wide level and air pollution forecast at monitoring stations. Section \ref{sec:experimental_setting_results} describes the experimental setting and results in detail. Section \ref{sec:discussions} discusses the implications of the experimental results and proposes future work directions. Finally, Section \ref{sec:conclusions} concludes this study.

\section{Related Work}\label{sec:related_work}

Our literature review below points out the strengths and limitations of existing air quality modeling studies and highlights the challenges of urban air quality modeling using deep learning.
\subsection{Physical-based Urban Air Quality Modeling}
Physical-based models have been proposed to simulate the air pollution process in urban areas characterized by complex building geometries and road networks \cite{lateb2016use}. Utilizing pollutant emissions from different sectors such as industry, household, and transportation, fine-grained air pollution estimation can be achieved by solving computational fluid dynamics (CFD) equations that describe the physical and chemical processes in urban environments \cite{kwak2015urban}. With simplified assumptions on the pollutant distribution, semi-empirical models such as the Gaussian dispersion model can reduce the computational complexity of CFD \cite{riddle2004comparisons}. Although physical-based models have capitalized on the scientific understanding of the pollution diffusion process, they have drawbacks, including the high computational cost \cite{mallet2018meta} and the inaccuracies and uncertainties in time-dependent inputs such as traffic emission estimates \cite{zhang2012real}. Such limitations have made it difficult for physical-based models to provide fine-grained air pollution estimation in a large geographical scale such as the entire city in real-time.
\subsection{Traditional Data-driven Urban Air Quality Modeling}
Data-driven approaches to urban air quality modeling, departing from physical-based models, are based on patterns learned from historical data. With some assumptions on the air pollution process, such models have achieved a lower computational cost and a comparable or better performance. Early attempts adopted statistical models for urban air quality modeling, including inverse distance weighting (IDW), Kriging, and land-use regression for fine-grained air pollution estimation \cite{briggs2000regression, wu2018hybrid}, and autoregressive integrated moving average (ARIMA) for air pollution forecast at monitoring stations \cite{kumar2004forecasting}. More advanced data-driven air quality modeling studies were undertaken based on machine learning models such as support vector regression (SVR), random forest (RF), and artificial neural network (ANN) \cite{rybarczyk2018machine}. These data-driven models capitalized on the strengths of machine learning in capturing the non-linear relationship between urban dynamics, including air quality and proxy data (i.e., auxiliary information such as meteorology). However, the number of air quality monitoring stations is often limited, making it challenging to train machine learning models given that ground-truth air pollution measurements are sparse.

Two major approaches have been adopted to tackle the data sparsity issue. On the one hand, to improve the coverage of real-time air quality measurements, portable sensors can be deployed to different urban environments via participatory sensing \cite{nikzad2012citisense}, vehicular sensing \cite{huang2018crowdsource}, or unmanned aerial vehicle (UAV) sensing \cite{yang2017real}. However, a large-scale sensor deployment throughout the city tends to be highly costly and requires significant effort for sensor calibration \cite{maag2018survey}. On the other hand, some advanced data-driven models have sought to better capture the spatial correlation of air pollution and proxy data for fine-grained air quality estimation at the city-wide level. These studies often divided a city into disjoint grids (e.g., 1km x 1km) and assumed that air pollution values in the same grid remain constant. Zheng et al. \cite{zheng2013u} proposed a semi-supervised machine learning method to estimate air pollution in grids not covered by monitoring stations by jointly training a spatial classifier (ANN) utilizing spatial features including POIs and road networks, and a temporal classifier (conditional random field (CRF)) using temporal features including meteorology, traffic, and human mobility. Along this line, Chen et al. \cite{chen2016spatially} proposed a semi-supervised ensemble learning model for air pollution estimation at a target grid, highlighting the importance of selecting spatial features from the nearest grids having monitoring data and sharing similar characteristics. Further, Zhu et al. \cite{zhu2017extended} proposed a Granger-causality-based data-driven model to estimate air pollution levels in a target grid, based on Granger-causal urban dynamics obtained from the most influential grids (which could be geographically far away). By selecting the most relevant urban dynamics data, the Granger-causality-based model achieved higher accuracy than baseline models using all urban dynamics data from nearby grids. Moreover, several advanced data-driven models have attempted to forecast air pollution at monitoring stations, while utilizing the spatial correlation of air pollution and proxy data. By modeling the spatial dependence and temporal dependence between air pollution and urban proxy data separately, Zheng et al. \cite{zheng2015forecasting} developed a hybrid machine learning framework consisting of a spatial predictor (ANN) and a temporal predictor (CRF) to forecast hourly air pollution levels at monitoring stations in the next two days. Zhao et al. \cite{zhao2017incorporating} proposed a multi-task learning framework to jointly estimate city-wide fine-grained air pollution in the current hour and forecast hourly air pollution at monitoring stations in the next three hours. Nevertheless, given that traditional machine learning models have not learned complex non-linearities from deep representations of spatio-temporal data \cite{wang2019deep}, it remains difficult for traditional data-driven models to model urban air quality accurately.

\subsection{Deep Learning-based Data-driven Urban Air Quality Modeling}
Deep learning or deep neural network models have advanced the state-of-the-art in data-driven urban air quality modeling. By learning deep representations and complex non-linear relationships from a large amount of heterogeneous spatio-temporal data in urban environments, deep learning models have achieved higher accuracy in air quality modeling tasks, including fine-grained air pollution estimation at the city-wide level and air pollution forecast at the monitoring station level.

On the one hand, a number of deep learning models have been proposed for fine-grained air pollution estimation at the city-wide level. Given that air quality measurements are usually geographically sparse, deep learning models, like other machine learning approaches that are often data-intensive, have faced significant challenges due to the lack of ground truths. On the one hand, vehicular sensing platforms have been developed to tackle the data sparsity issue. Ma et al. \cite{ma2020fine} proposed an autoencoder framework to recover a real-time high-resolution air pollution map covering a district in a city, based on a ConvLSTM model. Similarly, Do et al. \cite{do2020graph} proposed an autoencoder framework to recover real-time high-resolution air pollution in discrete locations in a city, based on a GCN model. On the other hand, without auxiliary sensors deployed, some deep learning studies, with a low deployment cost, have been proposed to better capture the spatio-temporal variation of air pollution across the city, utilizing urban proxy data that are already readily available. Cheng et al. \cite{cheng2018neural} proposed an attention-based hybrid deep learning framework based on the intuition that not all monitoring data contributed equally to air quality levels at a specific location. The attention model integrated an LSTM model for sequential data (air quality and meteorology) modeling and a feedforward neural network for spatial data (POIs and road networks) modeling to automatically learn the weights of air pollution monitoring stations for estimation in new locations in the current hour. As another line of research, Ma et al. \cite{ma2018guiding}  utilized a multi-task learning framework based on the observed air quality data and fine-grained air quality estimations generated by a dispersion model, highlighting the use of physical-based models to guide the neural network training process.

On the other hand, by exploiting the strengths of deep learning in modeling the non-linear temporal correlation of time series data, many studies have demonstrated better performance of deep learning models for air pollution forecast at monitoring stations. Earlier studies utilized the recurrent neural network (RNN) model \cite{ong2016dynamically} and its variants, including the LSTM and gated recurrent unit (GRU) model \cite{li2017long, athira2018deepairnet, deepair2017, freeman2018forecasting, zhou2019explore}, to capture the temporal dependence of air pollution and meteorology. More recent studies have extended the RNN-based air quality modeling using a variety of techniques, such as decomposing the air pollution time series into different frequencies of components \cite{jiang2020novel}, filling in missing proxy data via an iterative network training method \cite{li2017deep}, incorporating weather forecast in the sequence-to-sequence (Seq2Seq) modeling \cite{luo2019accuair}, focusing on the most relevant information using a spatial attention mechanism \cite{liu2020air, shi2020novel} or a temporal attention mechanism \cite{liu2018attention}, or accounting for forecast uncertainties using Bayesian methods \cite{han2020domain}. Moreover, a transfer learning framework using a bidirectional LSTM model was proposed to forecast air pollution at a newly built station where monitoring data was only collected for one month \cite{ma2020air}. However, these extensions failed to address to the spatial dependence of air pollution data. By incorporating a CNN or GCN component into the RNN-based model, hybrid deep learning models such as CNN-LSTM \cite{soh2018adaptive, qin2019novel, wen2019novel, zhang2020deep, pak2020deep, yan2020multi}, GCN-LSTM \cite{qi2019hybrid, ge2020multi}, and GCN-GRU \cite{lin2018exploiting, wang2020pm2} models were proposed for air pollution forecast, to take into account the spatial dependence of nearby observations including air quality and auxiliary data such as meteorology and urban morphology. In addition to the RNN-based modeling, the one-dimensional CNN (1D-CNN) model was used to extract the temporal dependence of urban dynamics observed at a station \cite{huang2018deep} or nearby stations \cite{du2019deep}. A deep fusion network consisting of multiple deep feedforward neural networks for air pollution forecast was proposed by capturing the complex interactions between different influential factors such as air pollutants and meteorological conditions \cite{yi2020predicting, yi2018deep}.

Moreover, a few deep learning models have been proposed to simultaneously provide fine-grained air pollution estimations for the entire city and air pollution forecasts for monitoring stations. Chen et al. \cite{chen2019deep} developed a multi-task CNN-LSTM framework to estimate fine-grained air pollution in the current hour and forecast air pollution at monitoring stations in the next 48 hours through shared spatio-temporal representations across different grids in a city. Nevertheless, the above deep learning studies did not fully consider important street canyon-related features and their spatial interactions, such as road density, building density/height, and street canyon. It remains an open question how advanced spatial models such as CNN models can be utilized to better account for important spatial features and their interactions relevant to the air pollution process in urban environments.

\subsection{Domain-specific Deep Learning for Urban Air Quality Modeling}
Although deep learning models have achieved state-of-the-art performance in air quality modeling, they tend to suffer from model overfitting due to limited and biased data. Moreover, the interpretability of deep learning models is often low, given their "black boxes" nature. Existing studies have highlighted the importance of domain-specific modeling to improve the generalizability and interpretability of deep learning-based air quality modeling. On the one hand, domain-specific auxiliary features that are highly relevant to the air pollution process have been taken into account, such as meteorology, weather \cite{yi2018deep}, POIs such as buildings \cite{lin2018exploiting} and factories \cite{liu2020air}, traffic conditions \cite{li2017deep}, road networks \cite{cheng2018neural}, factory emissions \cite{chen2019deep}, and time features such as month and day of the week \cite{yi2018deep}. On the other hand, the incorporation of domain-specific knowledge to guide the model learning process has been investigated. Han et al. \cite{han2020domain} incorporated a specific regularization term into the model training procedure to penalize PM predictions inconsistent with domain knowledge, particularly, the high correlation between PM\textsubscript{2.5} and PM\textsubscript{10} pollutants observed in empirical studies. Ma et al. \cite{ma2018guiding} utilized domain knowledge adopted in an air pollution dispersion model to calculate a neural network model's loss function based on the observed air quality data and the simulated pollution data generated by the dispersion model. Moreover, Ma et al. \cite{ma2020fine} interpreted the connection between a ConvLSTM model and a simplified dispersion model, demonstrating that the dispersion model's coefficients were automatically learned from data during the model training. Nevertheless, without proper changes in the model structure, domain-specific model training might still fail to address the complex interaction between various factors contributing to air quality changes over space and time.

Domain knowledge has been exploited in tailor-making deep learning model structures to better capture the air pollution process. The spatial and temporal interaction between important features, such as air pollution and meteorology, has been increasingly incorporated. Yi et al. \cite{yi2018deep} proposed an ensemble deep learning framework, where each component (deep feedforward network) was designed according to domain knowledge, i.e., the direct and indirect factors that can affect air quality. Liu et al. \cite{liu2020air} utilized an attention-based LSTM model to learn the impact of factory-related factors on local PM\textsubscript{2.5} concentrations. However, the deep feedforward network or LSTM model was incapable of addressing the complicated spatial relationship between those factors. Lin et al. \cite{lin2018exploiting} constructed a unidirectional graph based on the similarity between the monitoring locations and the nearby influential features such as roads and buildings, using a diffusion convolution method to extract the spatial dependence from the graph-structured data. Chen et al. \cite{chen2019deep} proposed a hybrid CNN-LSTM model for air quality prediction, utilizing a graph embedding layer to generate high-level representations of spatial data as inputs to a CNN model, while preserving the spatial relationship among nodes in the POI and road network graphs. Wang et al. \cite{wang2020pm2} proposed a knowledge-graph-based hybrid GCN-GRU model for PM\textsubscript{2.5} pollution forecast, where domain knowledge was explicitly encoded into a bidirectional graph as attributes of nodes (such as wind speed) and edges (such as the impact of wind speed on PM\textsubscript{2.5} transport from one node to another). However, until now, the street canyon effect has largely been overlooked in existing domain-specific deep learning studies. The spatial interaction between various urban dynamics has yet to be fully addressed by deep learning-based air quality models to capture the street-level variation of air pollution in urban areas characterized by high-rise buildings and complex traffic conditions.

\subsection{Research Gap}
In summary, it remains to be investigated in more detail how advanced deep learning models can better capture the characteristics of the complex spatial interaction among air pollution and urban proxy data, especially given that existing data-driven models have yet to address the street canyon effect explicitly. To the best of our knowledge, until now, no deep learning model has been proposed to take into account the street canyon effect in urban environments for fine-grained air pollution estimation at the city-wide level and air pollution forecast at monitoring stations (see Table \ref{table:related_work} for a summary of related work). Based on our previous work in urban air quality modeling \cite{zhang2020deep}, this study aims to fill this gap by proposing a hybrid CNN-LSTM model to capture the spatio-temporal correlation between air pollution and other important urban dynamics (e.g., meteorology, traffic speed, road density, building density/height, and street canyon), utilizing 1x1 convolution layers that facilitate the spatial information exchange across various urban dynamics.

\begin{table*}[htb]
    \centering
    \caption{A Summary of Related Deep Learning Studies in Urban Air Quality Modeling and The Added Value of This Study}
    \centering
    \begin{tabular}{|m{0.14\textwidth}|m{0.13\textwidth}|m{0.08\textwidth}|m{0.49\textwidth}|m{0.03\textwidth}|}
    \hline
    \textbf{Category} & \textbf{Model} & \multicolumn{2}{|l|}{\textbf{Domain-specific Modeling}} & \textbf{Ref.} \\
    \hline
    \multicolumn{5}{|l|}{\textbf{(a) Fine-grained air pollution estimation at the city-wide level}} \\
    \hline
    \multirow{2}{=}{Autoencoder modeling} & ConvLSTM & Features & Air pollution and meteorology (collected from portable sensors) & \cite{ma2020fine} \\
    \cline{2-5}
     & GCN & Features & Air pollution (collected from portable sensors) & \cite{do2020graph} \\
    \hline
    Attention modeling & LSTM-FNN & Features & Air pollution, meteorology, road network, POI & \cite{cheng2018neural} \\
    \hline
    \multirow{4}{=}{Physical-inspired learning} & \multirow{2}{*}{FNN} & Features & Air pollution (observed and simulated data) & \multirow{2}{*}{\cite{ma2018guiding}} \\
    \cline{3-4}
    &  & Training & Model training guided by predictions generated from an air pollution dispersion model & \\
    \cline{2-5}
    & \multirow{2}{*}{ConvLSTM} & Features & Air pollution and meteorology (collected from portable sensors) & \multirow{2}{*}{\cite{ma2020fine}} \\
    \cline{3-4}
    &  & Training & Model training guided by the connection between training a convolutional LSTM model and estimating parameters of a dispersion model & \\
    \hline
    
    \multicolumn{5}{|l|}{\textbf{(b) Air pollution forecast at the monitoring station level}} \\
    \hline
    \multirow{8}{=}{RNN-based modeling} & RNN (LSTM) & Features & Air pollution, meteorology, traffic condition & \cite{li2017deep} \\
    \cline{2-5}
    & RNN (GRU) & Features & Air pollution, meteorology & \cite{athira2018deepairnet} \\
    \cline{2-5}
    & Seq2Seq (LSTM) & Features & Air pollution, meteorology, weather forecast & \cite{liu2018attention} \\
    \cline{2-5}
    & Seq2Seq (GRU) & Features & Air pollution, meteorology, weather forecast & \cite{luo2019accuair} \\
    \cline{2-5}
    & Seq2Seq Attention & Features & Air pollution, meteorology, weather forecast & \cite{liu2018attention} \\
    \cline{2-5}
     & \multirow{2}{*}{Spatial Attention} & Features & Air pollution and factory-related features (e.g., location, land use, and product type) & \multirow{2}{*}{\cite{liu2020air}} \\
     \cline{3-4}
     &  & Structure & An attention layer to learn the impact of factory-related features on local PM\textsubscript{2.5} concentrations & \\
     \cline{2-5}
     & \multirow{2}{=}{Bayesian RNN} & Features & Air pollution, meteorology, weather forecast & \multirow{2}{*}{\cite{han2020domain}}\\
     \cline{3-4}
     &  & Training & Model training guided by the high correlation between PM\textsubscript{2.5} and PM\textsubscript{10} concentrations & \\
    \hline
    \multirow{11}{=}{Hybrid spatio-temporal modeling} & CNN-LSTM & Features & Air pollution, meteorology, planetary boundary layer height, aerosol optical depth & \cite{wen2019novel} \\
    \cline{2-5}
    & 1D-CNN-LSTM & Features & Air pollution, meteorology & \cite{du2019deep} \\
    \cline{2-5}
    & GCN-LSTM & Features & Air pollution, meteorology, station location, time features & \cite{qi2019hybrid} \\
    \cline{2-5}
    & \multirow{2}{*}{GCN-GRU} & Features & Air pollution, meteorology, geographic features (e.g., road length, land use, and building type) & \multirow{2}{*}{\cite{lin2018exploiting}} \\
    \cline{3-4}
     &  & Structure & Using a diffusion convolution method to extract the spatial relationship based on the important geographic features that can affect air quality &  \\
     \cline{2-5}
     & \multirow{2}{=}{GCN-GRU} & Features & Air pollution, meteorology, weather forecast, planetary boundary layer height & \multirow{2}{*}{\cite{wang2020pm2}} \\
     \cline{3-4}
     &  & Structure & Utilizing a knowledge graph capturing the interaction between air pollution and meteorology &  \\
    \hline
    \multirow{2}{=}{Ensemble learning} & \multirow{2}{*}{FNN} & Features & Air pollution, meteorology, weather forecast, time features, station ID & \multirow{2}{*}{\cite{yi2020predicting}} \\
    \cline{3-4}
     &  & Structure & Designing a set of FNN models based on the interactions of direct and indirect factors that can affect air quality &  \\
    \hline
    Transfer learning & LSTM & Features & Air pollution (existing stations and a new station) & \cite{ma2020air}\\
    \hline
    
    \multicolumn{5}{|l|}{\textbf{(c) Fine-grained air pollution estimation at the city-wide level and air pollution forecast at the monitoring station level}} \\
    \hline
    \multirow{4}{=}{Hybrid spatio-temporal modeling} & \multirow{2}{=}{Graph Embedding plus CNN-LSTM} & Features & Air pollution, meteorology, weather forecast, traffic condition, factory emission, road network, POI (e.g., factory) & \multirow{2}{*}{\cite{chen2019deep}} \\
    \cline{3-4}
     &  & Structure & Using a graph embedding layer to preserve the spatial relationship in the POI and road network graphs & \\
    \cline{2-5}
    & \multirow{2}{=}{\textbf{CNN-LSTM with 1x1 convolution layers (proposed model)}} & Features & Air pollution, meteorology, traffic condition, road density, \textbf{building density/height}, \textbf{street canyon}, time features & \multirow{2}{*}{N.A.} \\
    \cline{3-4}
     &  & Structure & \textbf{Using 1x1 convolution layers to strengthen the spatial representation learning for the interaction between air pollution, meteorology, traffic conditions, and urban morphology (road density, building density/height, and street canyon), to address the street canyon effect} & \\
    \hline
    \end{tabular}
    \label{table:related_work}
\end{table*}

\section{Methodology}\label{sec:method}
\begin{figure*}[!t]
    \centering
    \includegraphics[width=0.9\textwidth]{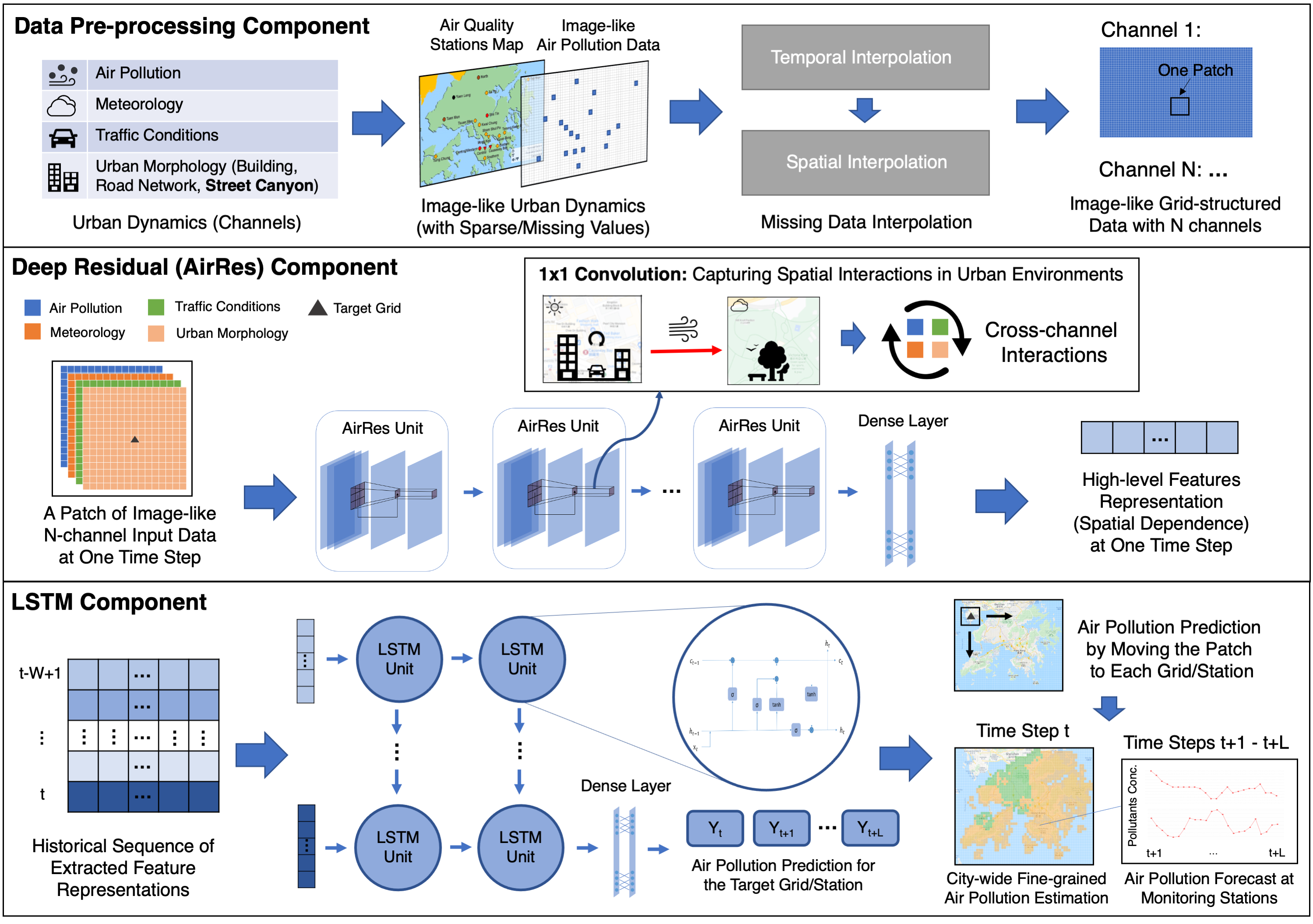}
    \caption{The Overview Structure of Deep-AIR, a Hybrid CNN-LSTM Deep Learning Framework}
    \label{fig:framework}
\end{figure*}

This study proposes Deep-AIR, a deep learning framework to estimate fine-grained air quality at the city-wide level in the current hour and forecast air pollution at monitoring stations up to 24 hours ahead, utilizing readily available urban dynamics data. The framework consists of three sequential components, including (1) a data pre-processing component to generate an image-like grid-structured dataset, with sparse/missing values interpolated temporally and spatially, (2) a residual CNN (AirRes) component for extracting spatial features and their interactions, using 1x1 convolution layers that facilitate the information exchange across different urban dynamic features, and (3) an LSTM component for modeling the temporal dependence of the extracted spatial representations for air pollution prediction. Figure \ref{fig:framework} shows the overall structure of Deep-AIR.

Under our proposed framework, two deep learning models were developed for fine-grained air pollution estimation and air pollution forecast separately, using the same network structure except the final output layer (see Sections \ref{subsec:residual_component} and \ref{subsec:lstm_component} for more detail). The key difference between these two models being the way in which the observed air pollution dynamics are used as the model input (see Section \ref{subsec:pre_processing} for more detail). For the fine-grained air pollution estimation model, air pollution dynamics at each monitoring station are interpolated by other stations, as the model targets at estimating air pollution levels in areas where air quality measurements are absent. In contrast, for the air pollution forecast model, air pollution dynamics at each monitoring station become the observed ground truths, as the model targets at forecasting air pollution levels at the monitoring station level where historical air quality measurements are available.

\subsection{Data Pre-processing Component}\label{subsec:pre_processing}
The air pollution and urban proxy data were pre-processed as follows. First, a city map was divided into thousands of disjoint grids by longitude and latitude. Each grid was associated with air pollution data and other urban dynamics data (including meteorology, traffic conditions, and urban morphology data) for every time step. As a result, the input data structure for the whole city was like a sequence of n-channel images. In the image-like data, each pixel corresponded to a grid on the map, and each channel corresponded to one kind of air pollution or other urban dynamics data. Two time-related features were also included to account for seasonal effects, corresponding to the season and workday/weekend labels. Second, a two-stage interpolation method was adopted for recovering missing values in the temporal dimension and spatial dimensions separately, in order to reduce the noise in the incomplete dataset and obtain a fixed input size for model training. In the first stage, a linear interpolation of historical data in the temporal dimension was conducted. In the second stage, a spatial interpolation for air pollution data and meteorological data was performed to impute each grid's missing values. The spatial interpolation method was based on the nearest observations, using the inverse distance squared weighting function. The squared weighting function was chosen (among the linear, squared, and cubic weighting functions) by trial and error.

The spatial interpolation procedure was different for fine-grained air pollution estimation at the city-wide level and air pollution forecast at monitoring stations. We removed the local air quality information when generating the grid-structured air pollution dynamics for fine-grained air pollution estimation. More specifically, we removed the observed air quality values for each air quality monitoring station and then filled in the missing values by interpolating observations at other stations. In contrast, we filled in the missing values in the grid-structured air pollution dynamics for forecasting air pollution at monitoring stations by interpolating the observed air quality values from all stations. Finally, after the temporal and spatial interpolations, a complete grid-structure dataset was generated for our proposed fine-grained air pollution estimation model and air pollution forecast model separately. 

\subsection{Deep Residual Component}\label{subsec:residual_component}
After data pre-processing, a sequence of city-wide "picture" of urban dynamics was obtained. The picture-like data was fed into a CNN model, capable of extracting spatial information from high-dimensional data. To better capture the complex spatial relationship between various urban dynamics, the structure of the CNN model was modified to improve (1) the capability in learning spatial representations through deeper network structures and (2) the information exchange across different channels (urban dynamics).

\subsubsection{Deep Residual Network}
Deep CNN models have achieved outstanding performance in learning high-level representations from spatial data. However, as the neural network layers continue to deepen, it becomes challenging to train the network model due to gradient exploding and gradient vanishing problems. Sometimes, adding more layers to a network model may even deteriorate the performance \cite{he2016deep}. A deep residual network (ResNet) model was proposed to overcome the gradient exploding/vanishing problems when training deeper neural networks \cite{he2016deep}. ResNet is a type of CNN that adds an identity mapping on each network block \cite{he2016deep}. A ResNet is made up of a series of blocks (residual units), and a residual unit consists of a few convolutional layers and an identity mapping, as shown in Equation \ref{eqn:residual}.

\begin{equation}
\label{eqn:residual}
X^{(l+1)}=X^{(l)} + \mathcal{F}(X^{(l)})
\end{equation}
where \(X^{(l)}\) and \(X^{(l+1)}\) denote the input and the output matrix of the \(l^{th}\) unit, respectively, and \(\mathcal{F}\) is the identity mapping function.
The residual units create a shortcut for the information flow, thus benefiting the training process of very deep networks. The capability of ResNet to capture spatial features through deeper networks has been demonstrated in other urban computing scenarios using spatio-temporal data, such as traffic flow prediction \cite{zhang2017deep}.

CNN model with a deep structure is needed to extract high-level representations from the spatial correlation between different urban dynamics for urban air quality modeling. Therefore, a ResNet model was utilized in our framework, but with modifications to better address the spatial interaction between different urban dynamics (see Section \ref{subsubsec:1x1_convolution} for more detail). The deep residual component was constructed using a series of modified residual units, processing the n-channel grid-structure input data at each time step, and mapping them into a sequence of feature vectors, representing the extracted spatial information from urban dynamics data.

\subsubsection{1x1 Convolution}\label{subsubsec:1x1_convolution}
\begin{figure}[!t]
    \centering
    \includegraphics[width=0.45\textwidth]{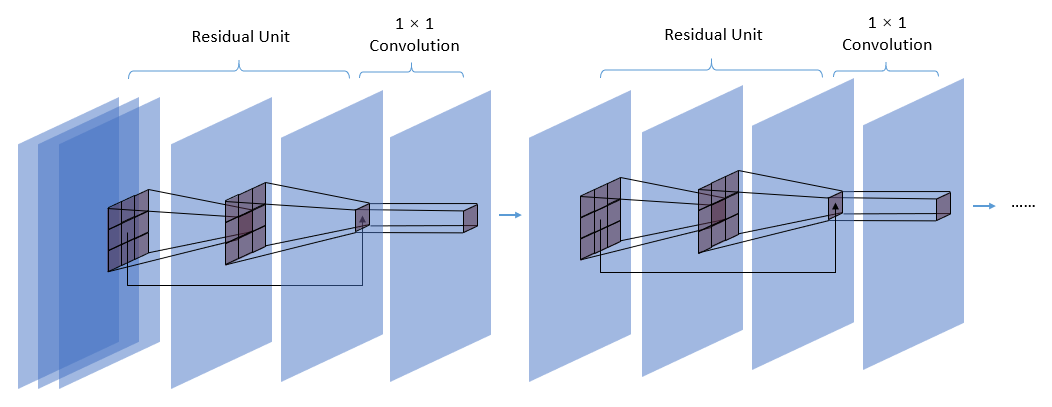}
    \caption{The Structure of the Proposed AirRes Model}
    \label{fig:airres_component}
\end{figure}

Although the image-like urban dynamics input data can be readily utilized by the ResNet model adopted in Deep-AIR, the unique characteristics of the air pollution process, particularly the cross-feature spatial interaction across air pollution and important urban dynamics, are yet to be fully taken into account. As mentioned in Section \ref{sec:introduction} and Figure \ref{fig:illstration}, the dispersion of air pollutants in urban environments is strongly dependent on influential factors such as meteorology (e.g., wind speed and direction can affect the transport of PM\textsubscript{2.5} pollutants) and street canyons (e.g., the levels of traffic pollution tend to be higher in high-density areas than open areas). Therefore, the ResNet model needs to be modified to strengthen the information exchange of different urban dynamic input channels. We developed a tailored ResNet model for air quality modeling, named AirRes, to address this challenge (see Figure \ref{fig:airres_component}). In the modified ResNet model, a 1x1 convolutional layer was inserted between each two adjacent residual units. 1x1 convolution is widely known for reducing the number of channels in GoogLeNet architecture \cite{szegedy2015going}. However, it can also facilitate information interflow across channels \cite{lin2013network} because the output of a 1x1 convolutional layer is equivalent to a linear combination of different feature maps.

\subsection{LSTM Component}\label{subsec:lstm_component}
\begin{figure}[!t]
    \centering
    \includegraphics[width=0.45\textwidth]{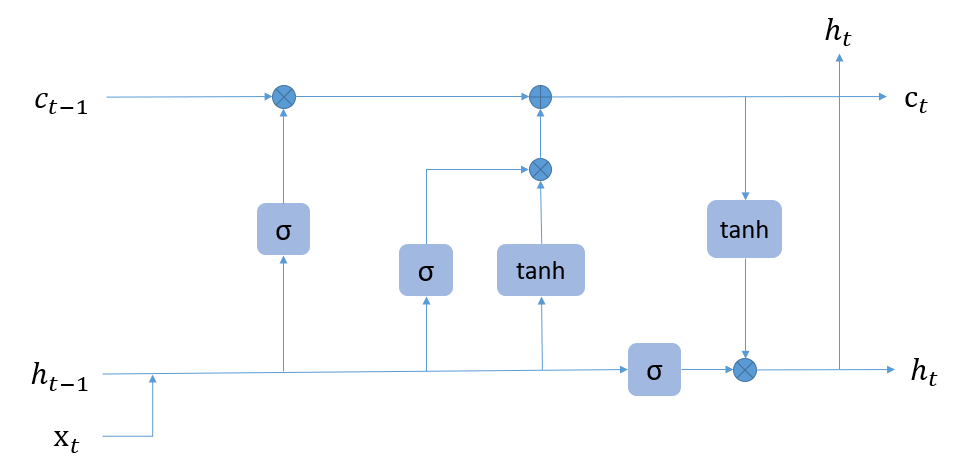}
    \caption{The Structure of an LSTM Block}
    \label{fig:lstm}
\end{figure}

An LSTM model is a special kind of RNN model characterized by advanced memory blocks rather than simple neurons at each time step. An LSTM's memory block consists of three gates to control the information flow within the memory block, namely, an input gate, a forget gate, and an output gate. Figure \ref{fig:lstm} shows the structure of an LSTM block. Due to the carefully designed gates, LSTM networks can avoid the gradient exploding/vanishing problem in RNN while remembering the long-term temporal correlation of sequential features, making it better in modeling time series data.

After extracting high-level spatial features through the deep residual component for each time step, the extracted feature matrix (a sequence of feature vectors for all time steps) was fed into the LSTM component. An LSTM's memory block is defined in Equation \ref{eqn:lstm}. The final hidden state of the LSTM component was used for air pollution prediction, using a fully connected layer. For fine-grained air pollution estimation at the city-wide level, one output was generated for an input matrix, representing the air quality in a grid in the current hour. For air pollution forecast at monitoring stations, multiple outputs were generated for an input matrix, representing the hourly air pollution levels at a monitoring station in the next hours.

\begin{equation}
\begin{aligned}
\label{eqn:lstm}
&\mathbf{i}_t = \sigma(\mathbf{W}_i \mathbf{x}_t + \mathbf{U}_i \mathbf{h}_{t-1}) \\
&\mathbf{f}_t = \sigma(\mathbf{W}_f \mathbf{x}_t + \mathbf{U}_f \mathbf{h}_{t-1}) \\
&\mathbf{o}_t = \sigma(\mathbf{W}_o \mathbf{x}_t + \mathbf{U}_o \mathbf{h}_{t-1}) \\
&\mathbf{c}_t = \mathbf{f}_t \odot \mathbf{c}_t + \mathbf{i}_t \odot \tanh(\mathbf{W}_c \mathbf{x}_t + \mathbf{b}_c)  \\
&\mathbf{h}_t = \tanh(\mathbf{c}_t) \odot \mathbf{o}_t
\end{aligned}
\end{equation}
where $t$ denotes a time step, $\mathbf{W}$ and $\mathbf{U}$ are the network weight and bias parameters, respectively, $\mathbf{x}_t$ is the extracted feature vector at the time step $t$, $\mathbf{h}_t$ is the hidden state at the time step $t$, $\mathbf{i}_t$, $\mathbf{f}_t$, $\mathbf{o}_t$ are the input, the forget, and the output gate at the time step $t$, respectively, $\mathbf{c}_t$ is the cell unit at the time step $t$, and $\odot$ denotes element-wise product.

\begin{algorithm}[t]
\caption{Patch Training for Deep-AIR}
\hspace*{-0.2\algorithmicindent} \textbf{Require:} \\
\hspace*{\algorithmicindent} input length $W$, forecasting period length $L$ ($L \geq 1$), \\
\hspace*{\algorithmicindent} estimation network model $f^{1}$ with initial parameters $\theta$, \\
\hspace*{\algorithmicindent} forecast network model $f^{2}$ with initial parameters $\phi$, \\
\hspace*{\algorithmicindent} patch size $N$, learning rate $\lambda$, \\
\hspace*{\algorithmicindent} the period of historical data $T$, stations $S$, grids $G$, \\
\hspace*{\algorithmicindent} air pollution measurements $\mathbb{P}=\{p_{s}^{t}\}_{t \in T, s \in S}$, \\
\hspace*{\algorithmicindent} air pollution dynamics $\mathbb{\hat{Q}}=\{\hat{q}_{g}^{t}\}_{t \in T, g \in G}$ (where values \\
\hspace*{\algorithmicindent} at each station are interpolated by other stations), \\
\hspace*{\algorithmicindent} air pollution dynamics $\mathbb{Q}=\{q_{g}^{t}\}_{t \in T, g \in G}$ (where values \\
\hspace*{\algorithmicindent} at each station are the observed ground truths),\\
\hspace*{\algorithmicindent} urban proxy dynamics $\mathbb{R}=\{r_{g}^{t}\}_{t \in T, g \in G}$, \\
\hspace*{\algorithmicindent} matrix concatenation $\oplus$
\begin{algorithmic}[1]
\STATE \textbf{function} \textsc{Train}($N$, $W$, $L$, $T$, $S$, $G$, $\mathbb{P}$, $\mathbb{Q}$, $\mathbb{\hat{Q}}$, $\mathbb{R}$, $\lambda$)
\REPEAT
    \FOR{$t$ from 1 to $T$}
        \STATE Sample a station $s$ from 1 to $S$
        \STATE Obtain $N \times N$ grids $\hat{G}$, with $\hat{G}$ centered on $s$
        \FOR{$t' \in [t-W+1,t]$}
            \STATE Create a patch $\hat{d}_{s}^{t'} = \{ \hat{q}_{k}^{t'} \oplus {r}_{k}^{t'} \}_{k \in \hat{G}}$ 
        \ENDFOR
        \STATE Estimate the current air pollution at station $s$: \\ $y_{s}^{t}=f^{1}_{\theta}(\hat{d}_{s}^{t-W+1},...,\hat{d}_{s}^{t})$
        \STATE Calculate the loss $\mathcal{L}=\|p_{s}^{t}-y_{s}^{t}\|^{2}$
        \STATE Perform back-propagation: \(\theta \gets \theta - \lambda\partial\mathcal{L}/\partial\theta\)
    \ENDFOR
\UNTIL{stopping criteria is met}
\REPEAT
    \FOR{$t$ from 1 to $T$}
        \STATE Sample a station $s$ from 1 to $S$
        \STATE Obtain $N \times N$ grids $\hat{G}$, with $\hat{G}$ centered on $s$
        \FOR{$t' \in [t-W+1,t]$}
            \STATE Create a patch $d_{s}^{t'} = \{ q_{k}^{t'} \oplus r_{k}^{t'} \}_{k \in \hat{G}}$ 
        \ENDFOR
        \STATE Forecast air pollution at station $s$: \\ $[y_{s}^{t+1},...,y_{s}^{t+L}]=f^{2}_{\phi}(d_{s}^{t-W+1},...,d_{s}^{t})$
        \STATE Calculate the loss $\mathcal{L}=\sum_{t+1}^{t+L}\|p_{s}^{t}-y_{s}^{t}\|^{2}/L$
        \STATE Perform back-propagation: \(\phi \gets \phi - \lambda\partial\mathcal{L}/\partial\phi\)
    \ENDFOR
\UNTIL{stopping criteria is met}
\RETURN fitted network models $f^{1}_{\theta}$ and $f^{2}_{\phi}$
\end{algorithmic}
\label{alg:training}
\end{algorithm}

\subsection{Model Training and Prediction}
The details of training the proposed Deep-AIR framework are shown as follows (see Algorithm \ref{alg:training}). After data pre-processing, a patch training algorithm was used to train two models for fine-grained air pollution estimation at the city-wide level and air pollution forecast at monitoring stations separately. During model training, for each pair of the input and output data, the input was a patch of the grid-structured map, with one monitoring station located at the center of the patch, and the output was the air pollution value measured at the center.

The details of estimating fine-grained air pollution and forecasting air pollution using our proposed Deep-AIR framework are as follows (see Algorithm \ref{alg:prediction}). For fine-grained air pollution estimation, a patch of historical urban dynamics data was generated for each grid on the map, including areas not covered by air quality monitoring stations. For air pollution forecast, a patch of historical urban dynamics data was generated for each air quality monitoring station on the map. Using the generated patches as the model inputs, the fine-grained estimation model predicted a fine-grained air pollution estimation map for the entire city in the current hour, and the air pollution forecast model predicted air pollution for each air quality monitoring station in the next hours.

\begin{algorithm}[t]
\caption{Fine-grained Air Pollution Estimation and Station-level Air Pollution Forecast}
\hspace*{-0.2\algorithmicindent} \textbf{Require:} \\
\hspace*{-0.2\algorithmicindent} the current time $t^{\star}$, the recent urban dynamics \hspace*{-0.2\algorithmicindent} $\mathbb{Q}$, $\mathbb{\hat{Q}}$, $\mathbb{R}$, \\
fitted $f^{1}_{\theta}$ and $f^{2}_{\phi}$, $N$, $W$, $L$, $S$, $G$
\begin{algorithmic}[1]
\STATE \textbf{function} \textsc{Predict}($t^{\star}$, $f^{1}_{\theta}$, $f^{2}_{\phi}$, $N$, $W$, $L$, $S$, $G$, $\mathbb{Q}$, $\mathbb{\hat{Q}}$, $\mathbb{R}$)
\FOR{$g$ from 1 to $G$}
    \STATE Obtain $N \times N$ grids $\hat{G}$, with $\hat{G}$ centered on $g$
    \FOR{$t' \in [t^{\star}-W+1,t^{\star}]$}
        \STATE Create a patch $\hat{d}_{g}^{t'} = \{ \hat{q}_{k}^{t'} \oplus {r}_{k}^{t'} \}_{k \in \hat{G}}$
        \IF{$g \in S$}
            \STATE Create a patch $d_{g}^{t'} = \{ q_{k}^{t'} \oplus {r}_{k}^{t'} \}_{k \in \hat{G}}$
        \ENDIF
    \ENDFOR
    \STATE Estimate the current air pollution at grid $g$: \\ $y_{g}^{t^{\star}}=f^{1}_{\theta}(\hat{d}_{g}^{t^{\star}-W+1},...,\hat{d}_{g}^{t^{\star}})$
    \IF{$g \in S$}
        \STATE Forecast air pollution at grid $g$: \\ $[y_{g}^{t^{\star}+1},...,y_{g}^{t^{\star}+L}]=f^{2}_{\phi}(d_{g}^{t^{\star}-W+1},...,d_{g}^{t^{\star}})$
    \ENDIF
\ENDFOR
\RETURN fine-grained air pollution map at $t^{\star}$ and station-level air pollution forecasts from $t^{\star}+1$ to $t^{\star}+L$
\end{algorithmic}
\label{alg:prediction}
\end{algorithm}

\section{Experimental Setting and Results}\label{sec:experimental_setting_results}

\subsection{Experimental Setting}
We collected two datasets, namely, Hong Kong for 16 months (Dec 2018-Mar 2020) and Beijing for 19 months (Jan 2017-Jul 2018). Table \ref{table:dataset} shows the data types and the corresponding data sources. We collected data in Hong Kong as follows. We collected the air pollution data from 16 public air quality monitoring stations from the Environmental Protection Department of the HKSAR Government (HKEPD). We collected the meteorology data from the Hong Kong Observatory (HKO). We downloaded the traffic conditions data from the Transport Department of the HKSAR Government (HKTD), covering traffic information on 617 major roads in Hong Kong. We obtained the information on roads and buildings in Hong Kong from the Lands Department of the HKSAR Government (HKLD). Moreover, our data collection procedure in Beijing is as follows. We collected the air pollution data from 35 public air quality monitoring stations from the Beijing Municipal Environment Monitoring Center (BJMEMC). We downloaded the meteorology data from 18 meteorology monitoring stations from the China Meteorological Data Service Center (CMDSC). We downloaded the traffic conditions data via the API provided by Gaode Map, covering traffic information on 227 major roads in Beijing. Urban morphology data were not available in Beijing. After data collection, for the data updated more frequently than once per hour, we averaged them for each hour so that the frequency of every kind of data was aligned to one hour.

\begin{table*}[htb]
    \caption{Urban Big Data Collected in Hong Kong from December 2018 to March 2020 and Beijing from January 2017 to July 2018}
    \centering
    \vspace*{0.5cm}
    \begin{tabular}{|l|l|l|l|l|l|}
    \hline
    \textbf{Domain}  & \begin{tabular}[c]{@{}l@{}}\textbf{Available}\\ \textbf{Points} \end{tabular}    &   \begin{tabular}[c]{@{}l@{}}\textbf{Update}\\ \textbf{Frequency}\end{tabular}  & \textbf{Data Category} & \textbf{Definition}  & \textbf{Data Source}       \\ \hline
    \multirow{5}{*}{\begin{tabular}[c]{@{}l@{}}Air \\ Pollution\end{tabular}} & \multirow{5}{*}{\begin{tabular}[c]{@{}l@{}}HK: 16* \\ BJ: 35\end{tabular}}  & \multirow{5}{*}{\begin{tabular}[c]{@{}l@{}}HK: 1hr\\ BJ: 1hr\end{tabular}}     & PM\textsubscript{2.5}   & PM\textsubscript{2.5} concentration ($\mu$g/$\mathrm{m}^3$)     & \multirow{6}{*}{\begin{tabular}[c]{@{}l@{}} HKEPD \cite{hkepd} \\ BJMEMC\cite{bjmemc}\end{tabular}} \\ \cline{4-5}
    &   &   & PM\textsubscript{10} & PM\textsubscript{10}  concentration ($\mu$g/$\mathrm{m}^3$)  &   \\ \cline{4-5}
    &   &   & NO\textsubscript{2} & NO\textsubscript{2} concentration ($\mu$g/$\mathrm{m}^3$)   &   \\ \cline{4-5}
    &   &   & SO\textsubscript{2}   & SO\textsubscript{2} concentration ($\mu$g/$\mathrm{m}^3$)   &   \\ \cline{4-5}
    &   &   & O\textsubscript{3}    & O\textsubscript{3} concentration ($\mu$g/$\mathrm{m}^3$)   &   \\ \cline{4-5}
    &   &   & CO**  & CO concentration ($\mu$g/m3)      &   \\ \hline
    \multirow{6}{*}{Meteorology}    & \multirow{6}{*}{\begin{tabular}[c]{@{}l@{}}HK: 44 \\ BJ: 18\end{tabular}}    & \multirow{6}{*}{\begin{tabular}[c]{@{}l@{}}HK: 10min \\ BJ: 1hr\end{tabular}}    & Pressure  & Atmospheric pressure (hPa)   & \multirow{6}{*}{\begin{tabular}[c]{@{}l@{}}HKO \cite{hko} \\ CMDSC \cite{cma} \end{tabular}}   \\ \cline{4-5}
    &   &   & Humidity  & Relative humidity (percentage)  &   \\ \cline{4-5}
    &   &   & Temperature   & Temperature (degree Celsius)    &   \\ \cline{4-5}
    &   &   & Wind speed    & Wind speed (km/h) &   \\ \cline{4-5}
    &   &   & \begin{tabular}[c]{@{}l@{}} Wind direction \end{tabular}  & \begin{tabular}[c]{@{}l@{}} Wind direction (eight possible values: \\E, W, N, S, NE, NW, SE, and SW) \end{tabular}  &   \\ \cline{4-5}
    &   &   & Precipitation*** & Precipitation (mm) &   \\ \hline
    \multirow{2}{*}{Traffic}    & \multirow{2}{*}{\begin{tabular}[c]{@{}l@{}}HK: 618 \\ BJ: 227\end{tabular}}    & \multirow{2}{*}{\begin{tabular}[c]{@{}l@{}}HK: 2min \\ BJ: 1hr\end{tabular}}    & \begin{tabular}[c]{@{}l@{}} Traffic congestion \end{tabular}  & \begin{tabular}[c]{@{}l@{}} Average congestion level of the \\road segmentation  within a grid \end{tabular} & \multirow{2}{*}{\begin{tabular}[c]{@{}l@{}}HKTD\cite{hktd} \\ Gaode\cite{gaode} \end{tabular}} \\ \cline{4-5}
    &   &   & \begin{tabular}[c]{@{}l@{}} Traffic speed \end{tabular}  & \begin{tabular}[c]{@{}l@{}} Average speed of the vehicles on the \\road segmentation within a grid \end{tabular}   &   \\ \hline
    \multirow{4}{*}{Morphology} &\multirow{4}{*}{\begin{tabular}[c]{@{}l@{}}HK: 44 * 60 \\ BJ: N.A.\end{tabular}} &\multirow{4}{*}{\begin{tabular}[c]{@{}l@{}}HK: as of 2018 \\ BJ: N.A.\end{tabular}} & Road density  & Number of road segments within a grid area  & \multirow{4}{*}{\begin{tabular}[c]{@{}l@{}}HKLD \cite{hkld} \end{tabular}} \\ \cline{4-5}
    &   &   & Building density  & Number of buildings within a grid area    &   \\ \cline{4-5}
    &   &   & Building height   & Average of the building heights within a grid area    &   \\ \cline{4-5}
    &   &   & \begin{tabular}[c]{@{}l@{}} Street canyon \end{tabular}  & \begin{tabular}[c]{@{}l@{}} Binary indicator (1: there is a street canyon effect in\\ this grid; 0: there is no street canyon effect in this grid) \end{tabular}  &   \\ \hline
    \multicolumn{6}{|l|}{\textbf{Notes:}}  \\
    \multicolumn{6}{|l|}{* Two newly built air quality monitoring stations in Hong Kong were not included during data collection.}  \\
    \multicolumn{6}{|l|}{** CO data in Hong Kong were not included because only some stations reported CO concentrations in Hong Kong.} \\
    \multicolumn{6}{|l|}{*** Precipitation data in Hong Kong were not provided by HKO.}  \\ \hline
    \end{tabular}
    \label{table:dataset}
\end{table*}

We pre-processed the two collected datasets as follows. We divided Hong Kong into 1km×1km grids, so the grid structure was an 18-channel 44×60 map (sixteen urban dynamics channels plus two time-label channels) for each time step. Similarly, we divided Beijing into 3km×3km grids, so the grid structure was a 16-channel 50×55 map (fourteen urban dynamics channels plus two time-label channels) for each time step. We generated two grid-structured datasets for each city and filled in missing values (see Section \ref{subsec:pre_processing} for more detail). The first grid-structured dataset was used for fine-grained air pollution estimation, whereas the second grid-structured dataset was used for air pollution forecast.

After data-processing, we conducted an experiment to train and evaluate our proposed model for fine-grained estimation (see Algorithm \ref{alg:training}). For Hong Kong and Beijing, we used a random 80/10/10 split of the first pre-processed grid-structured dataset as the training set, the validation set, and the testing set. Given that no local air pollution dynamics data were available for each grid, we obtained fine-grained air pollution estimations for all grids based on the interpolated air pollution dynamics and other urban proxy dynamics data. We also visualized the fine-grained PM\textsubscript{2.5} pollution estimates in Hong Kong and Beijing across four seasons to better understand the geographical and seasonal variation of the estimated levels of air pollution. Moreover, we conducted another experiment to train and evaluate our proposed model for forecasting one-hour and 24-hour air pollution at monitoring stations (see Algorithm \ref{alg:training}). For Hong Kong and Beijing, we used a random 80/10/10 split of the second pre-processed grid-structured dataset as the training set, the validation set, and the testing set. Given that local air pollution dynamics data were available for each station, we obtained air pollution forecasts at each air quality monitoring station based on the observed historical air pollution data and other urban proxy data.

We trained our proposed models, namely, the fine-grained air pollution estimation model and the air pollution forecast model, using the following settings. The patch size was set to 15. The AirRes component consisted of four residual units, and each residual unit had two 3x3 convolution layers with batch normalization and ReLU activation function. A 1x1 convolution layer was added between each two of the residual units. For the LSTM component, the length of the model input (past observations) was set to 48 (hours), the number of LSTM layers was set to one or two, and the hidden unit size was set to 128, 256, or 512. A stochastic gradient descent optimizer was used to train the model, and the learning rate of the optimizer was set to $10^{-4}$. The training process was stopped when the validation error was not improved in the latest five epochs. The best hyper-parameters, including the number of LSTM layers and the hidden unit size, were selected based on the performance evaluated on the validation set.

We included four baseline models for model evaluation and comparison. First, an autoregressive integrated moving average (ARIMA) model, a statistical method for time series analysis, was selected. Second, a standard LSTM model, a widely used method for deep learning-based air pollution modeling, was developed. Third, a standard ConvLSTM model, a typical structure for spatio-temporal data processing that integrates convolutional structures into an LSTM model, was constructed. Moreover, we selected a baseline model (ResNet-LSTM), using our proposed model without changing the ResNet structure into the AirRes structure (i.e., without using 1x1 convolution layers). We used the mean absolute percentage error (MAPE) as the performance metric (see Equation \ref{eqn:mape}).

\begin{equation}
\label{eqn:mape}
\mathrm{MAPE}=\sum_{i=0}^{n}\sum_{t=0}^{t=L}\dfrac{|y_{i}^{t}-{y^{t}_{i}}^{\star}|}{y_{i}^{t}} \times 100\%
\end{equation}
where $n$ is the sample size, $L$ is the length of the prediction period ($L$=0 for fine-grained air pollution estimation), $y_{i}^{t}$ is the ground truth air pollution value of the $i^{\mathrm{th}}$ test sample at time step $t$, and ${y^{t}_{i}}^{\star}$ is the predicted air pollution value of the $i^{\mathrm{th}}$ test sample at time step $t$.

In addition to performance evaluation, we conducted a saliency score analysis to better understand the influential features that can affect air quality prediction. As illustrated by the previous work \cite{simonyan2013deep}, the gradients with respect to the input values can reflect how much each input feature contributes to the output value. The output (i.e., air pollution prediction) near a single point can be approximately expressed by Equation \ref{eqn:grad}. The magnitude of each dimension of the gradient indicates the sensitiveness of the output values to that particular input feature. By taking the average of the absolute gradient for each model input over the whole training set, the saliency score is defined by Equation \ref{eqn:saliency}.

\begin{subequations}
\begin{equation}
\label{eqn:grad}
y \approx w(\mathbf{x})^T \mathbf{x} + b
\end{equation}
\begin{equation}
\label{eqn:saliency}
\mathbf{s}=\sum_{\mathbf{x},y \in D} \dfrac{\lvert w(\mathbf{x}) \rvert}{\lvert D \rvert}
\end{equation}
\end{subequations}
where $\mathbf{x}$ is the input data consisting of different features (urban dynamics), $y$ is the predicted air pollution value, $w$ and $b$ are respectively the weight parameters and the bias parameter to approximate the relationship between $\mathbf{x}$ and $y$ ($w$ is the derivative of $y$ with respect to $\mathbf{x}$), $D$ is the whole training set, and $\mathbf{s}$ is the saliency score vector that corresponds to the input features.

\subsection{Results}
We tested our proposed models and the baseline models based on the Hong Kong and Beijing datasets. Results show that our proposed Deep-AIR framework has achieved the best performance in fine-grained air pollution estimation and air pollution forecast, while providing interpretations on which features are most important in predicting urban air quality. We present the detailed results in the remaining parts of this section. We will discuss why our proposed model works better than the baseline models and which parts can be improved in Section \ref{sec:discussions}.

Our proposed model has shown better performance than the baseline models when making city-wide fine-grained air pollution estimations in the current hour. Table \ref{table:air_pollution_estimation} shows the average error rate of different models for fine-grained air pollution estimation at air quality stations when the local air pollution information was removed. The results show that our proposed model has achieved the lowest error compared to the baseline models. On average, the error rate of our proposed fine-grained estimation model was 32.4\% and 35.0\% in Hong Kong and Beijing, respectively.

\begin{table}[htb]
    \centering
    \caption{Error Rates (MAPEs) of Fine-grained Air Pollution Estimation in the Current Hour After Removing the Local Air Pollution Information}
    \begin{tabular}{|l|p{0.1\linewidth}|p{0.1\linewidth}|p{0.1\linewidth}|p{0.1\linewidth}|p{0.1\linewidth}|}
    \hline
    \textbf{City / Model} & ARIMA & LSTM & Conv-LSTM & ResNet-LSTM & Deep-AIR \\ \hline
    Hong Kong & 35.0  & 33.1 & 33.3 & 33.0 & \textbf{32.4} \\ \hline
    Beijing & 38.9  & 36.4 & 36.0 & 35.5 & \textbf{35.0} \\ \hline
    \end{tabular}
    \label{table:air_pollution_estimation}
\end{table}

\begin{table*}[htb]
    \centering
    \caption{Error Rates (MAPEs) of 1-hour and 24-hour Air Pollution Forecast in Hong Kong}
    \begin{tabular}{|l|l|l|l|l|l|l|l|l|l|l|}
    \hline
    \textbf{Model} & \multicolumn{2}{|c|}{ARIMA} & \multicolumn{2}{|c|}{LSTM} & \multicolumn{2}{|c|}{ConvLSTM} & \multicolumn{2}{|c|}{ResNet-LSTM} & \multicolumn{2}{|c|}{Deep-AIR} \\ \hline
    \textbf{Features / Period} & 1hr & 24hrs & 1hr & 24hrs & 1hr & 24hrs & 1hr & 24hrs & 1hr & 24hrs \\ \hline
    Air & 29.8 & 45.5 & 28.1 & 41.4 & 28.1 & 41.5 & 27.9 & 40.9 & 27.5 & 40.9 \\ \hline
    Air + Weather & 27.5 & 43.7 & 26.0 & 39.5 & 26.4 & 39.4 & 26.1 & 39.0 & 25.2 & 37.1 \\ \hline
    Air + Weather + Traffic & 26.2 & 42.3 & 25.1 & 38.1 & 25.0 & 37.9 & 24.8 & 37.5 & 24.3 & 35.1 \\ \hline
    Air + Weather + Traffic + Morphology & 26.0 & 42.3 & 24.7 & 37.5 & 24.2 & 37.0 & 24.0 & 35.5 & \textbf{22.8} & \textbf{33.9} \\ \hline
    \end{tabular}
    \label{table:air_pollution_forecast_hk}
\end{table*}

\begin{table*}[htb]
    \centering
    \caption{Error Rates (MAPEs) of 1-hour and 24-hour Air Pollution Forecast in Beijing}
    \begin{tabular}{|l|l|l|l|l|l|l|l|l|l|l|}
    \hline
    \textbf{Model} & \multicolumn{2}{|c|}{ARIMA} & \multicolumn{2}{|c|}{LSTM} & \multicolumn{2}{|c|}{ConvLSTM} & \multicolumn{2}{|c|}{ResNet-LSTM} & \multicolumn{2}{|c|}{Deep-AIR} \\ \hline
    \textbf{Features / Period} & 1hr & 24hrs & 1hr & 24hrs & 1hr & 24hrs & 1hr & 24hrs & 1hr & 24hrs \\ \hline
    Air & 32.8 & 54.5 & 31.0 & 44.5 & 30.6 & 43.7 & 28.7 & 42.1 & 28.5 & 42.2 \\ \hline
    Air + Weather & 31.9 & 53.2 & 29.6 & 42.4 & 28.8 & 41.1 & 27.4 & 39.9 & 26.3 & 39.2 \\ \hline
    Air + Weather + Traffic & 29.5 & 52.5 & 28.4 & 40.5 & 27.5 & 39.3 & 26.5 & 38.4 & \textbf{24.7} & \textbf{36.5} \\ \hline
    \end{tabular}
    \label{table:air_pollution_forecast_beijing}
\end{table*}

\begin{figure*} 
  \centering
  [a] \includegraphics[width=0.45\linewidth]{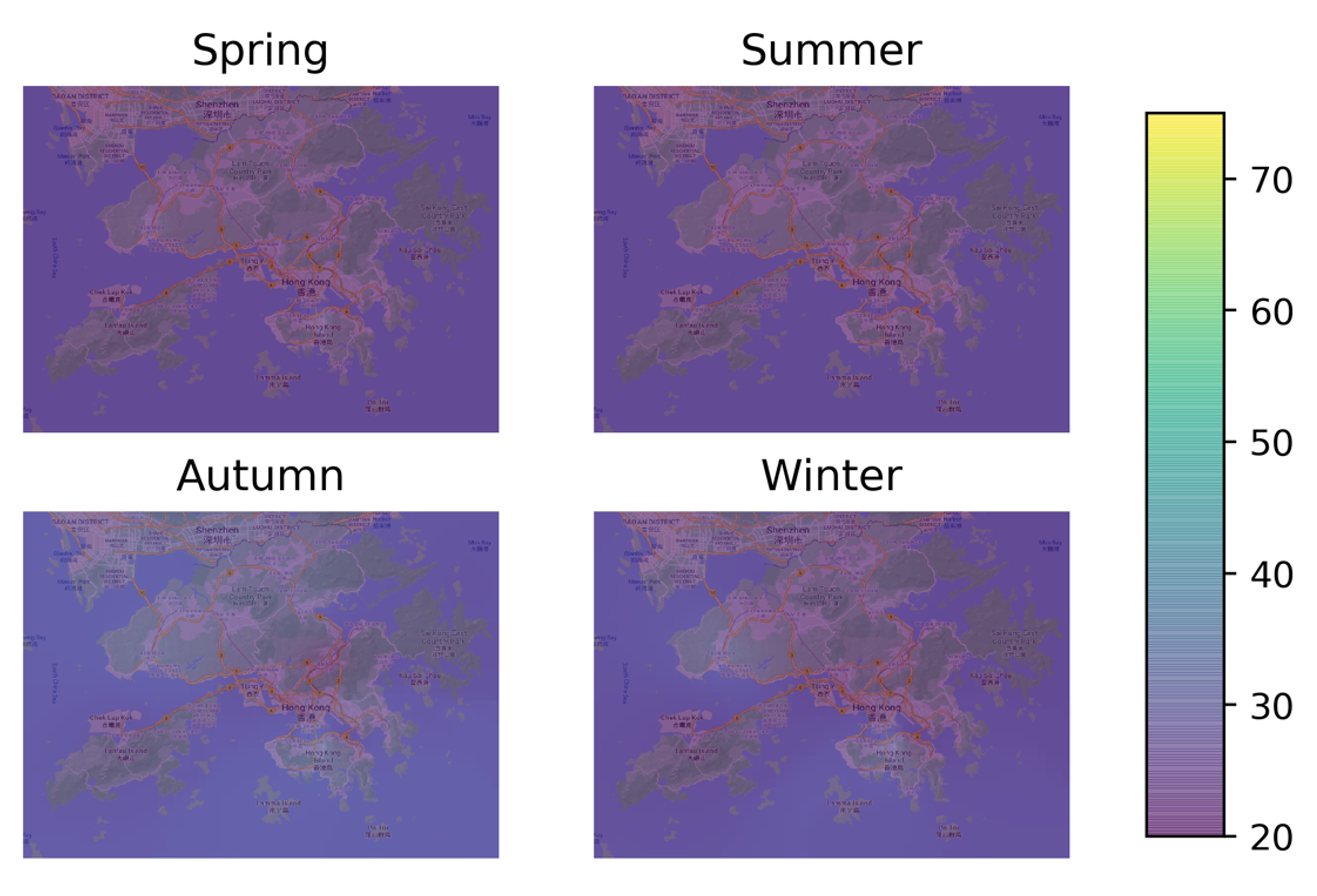}
  \hfill
  [b] \includegraphics[width=0.45\linewidth]{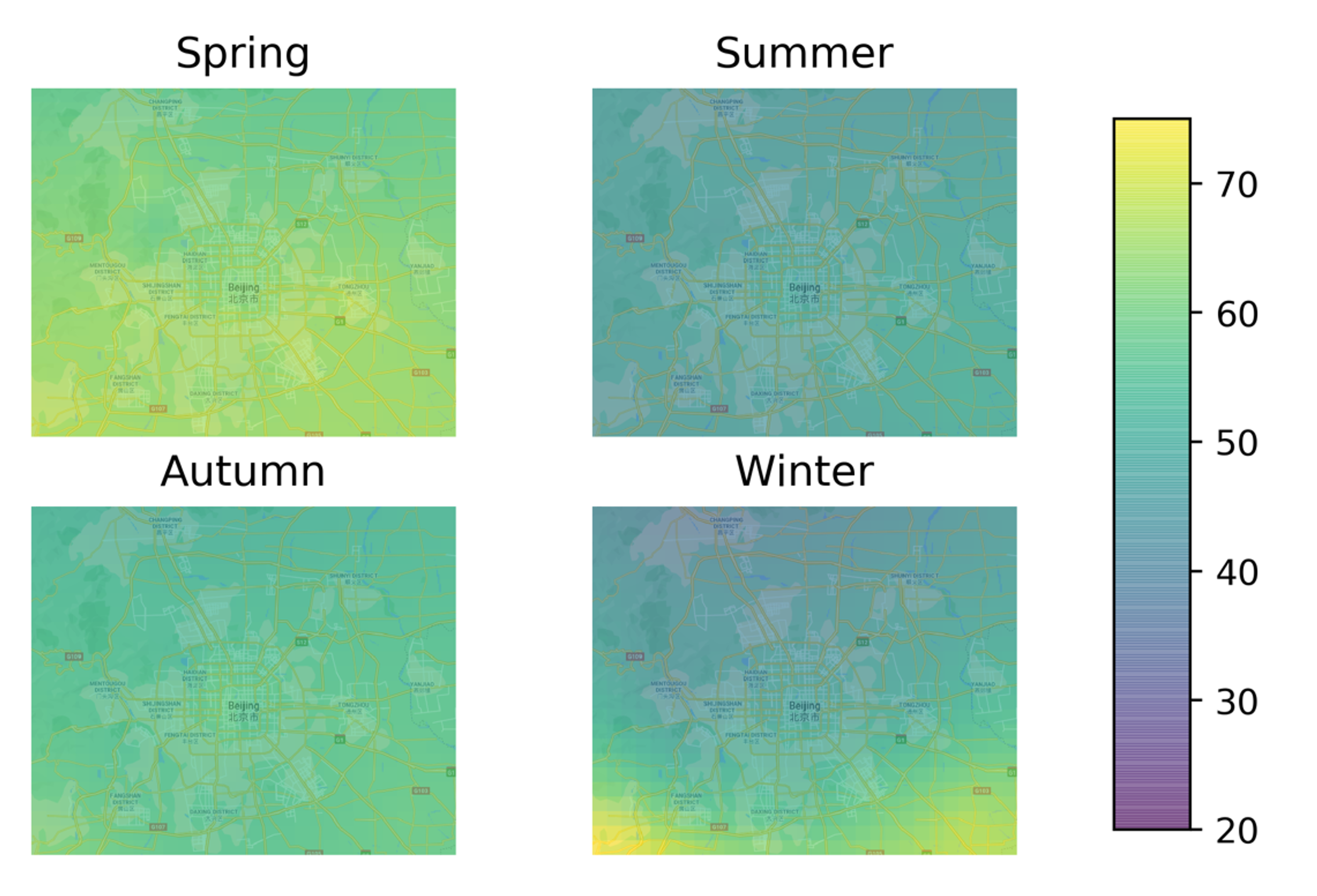}
  \caption{Comparison of the average PM\textsubscript{2.5} concentrations across four seasons in (a) Hong Kong (from March 2019 to March 2020) and (b) Beijing  (from March 2017 to March 2018)}
  \label{fig:pollution_visualization} 
\end{figure*}

Our proposed fine-grained estimation model has made possible the evaluation of air pollution level at a fine-grained scale throughout the city. The city-wide fine-grained estimation results are used to visualize the seasonal and geographical patterns of PM\textsubscript{2.5} pollution in Hong Kong and Beijing. Figure \ref{fig:pollution_visualization} shows the average PM\textsubscript{2.5} estimated values in Hong Kong and Beijing in spring (from March to May), summer (from June to August), autumn (from September to November), and winter (from December to February). The average PM\textsubscript{2.5} level was higher in Beijing than in Hong Kong during the study period. The two air quality visualization maps demonstrate seasonal variation in PM\textsubscript{2.5} estimated values. For example, the PM\textsubscript{2.5} estimates were slightly higher in winter and spring but lower in summer, especially in Beijing, which can be attributable to the heating supply. Moreover, the two maps illustrate the geographical variations of PM\textsubscript{2.5} pollution estimates. For example, the PM\textsubscript{2.5}  estimates were higher in the southern part of Beijing than in other parts of the city during winter, which can be attributable to the regional transportation of air pollutants across the Beijing-Tianjin-Hebei region. Nevertheless, the geographical variation of air pollution levels was small, especially in Hong Kong, probably because air pollution episodes became insignificant when averaging the hourly PM\textsubscript{2.5} estimates of individual seasons.

Moreover, our proposed model has outperformed the baselines for forecasting air pollution at monitoring stations in the next hour and over the next 24 hours. Tables \ref{table:air_pollution_forecast_hk} and \ref{table:air_pollution_forecast_beijing} show the average forecasting error rate of our proposed model and the baselines. Although forecasting air pollution levels for the next 24 hours is less accurate for all models, our proposed model achieves the lowest error compared to the baseline models, utilizing all available urban dynamics data. In Hong Kong, the lowest forecast error is 22.8\% in the next hour and 33.9\% over the next 24 hours. In Beijing, the lowest forecast error is 24.7\% in the next hour and 36.5\% over the next 24 hours. Moreover, when evaluating different sets of input features, all models have achieved better results when more types of features are inputted and poorer results when less types of features are inputted. When two or more types of urban dynamics are included, our proposed model has outperformed the baseline models.

\begin{figure*} 
  \centering
  [a]\includegraphics[width=0.45\linewidth]{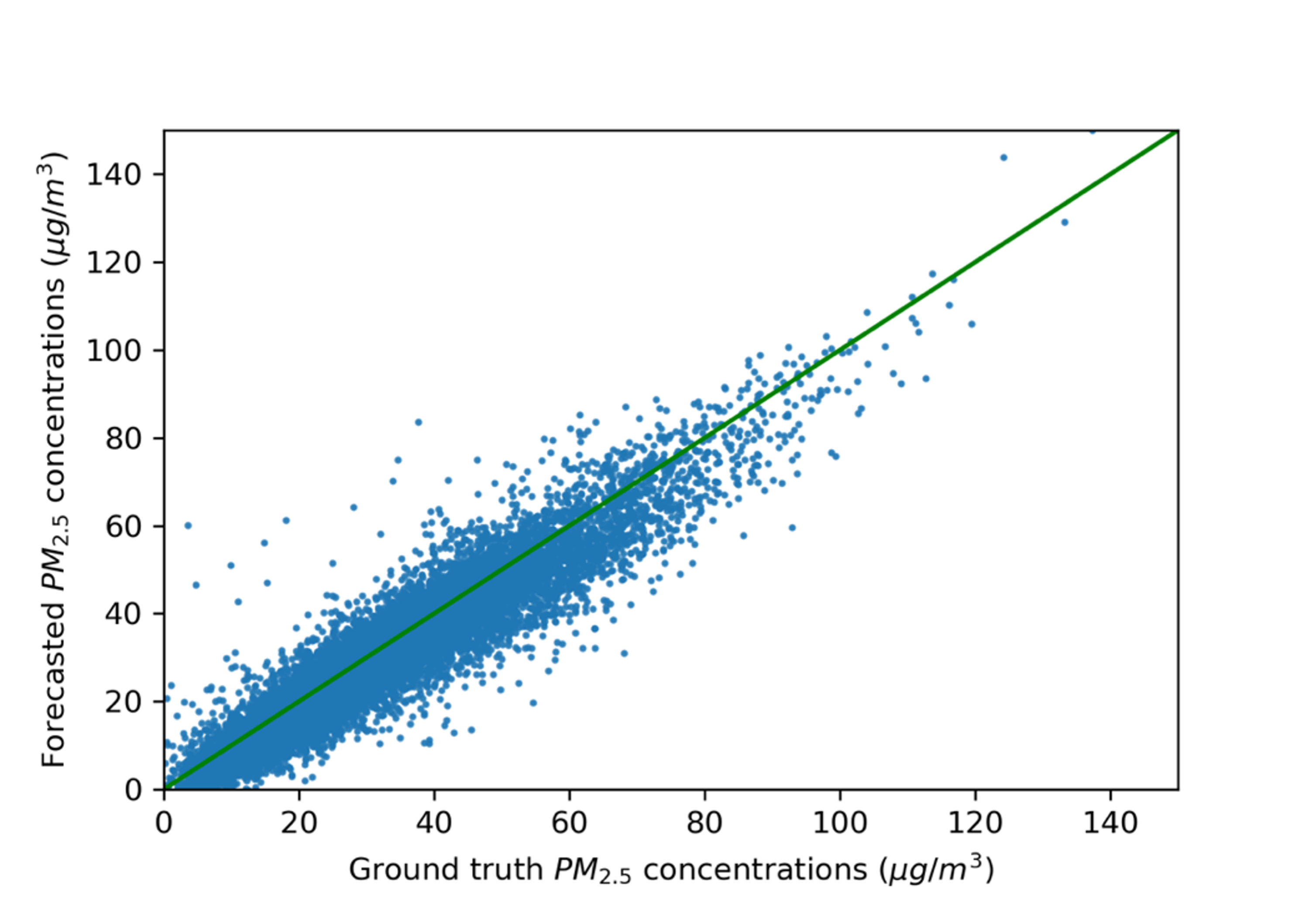}
  \hfill
  [b]\includegraphics[width=0.45\linewidth]{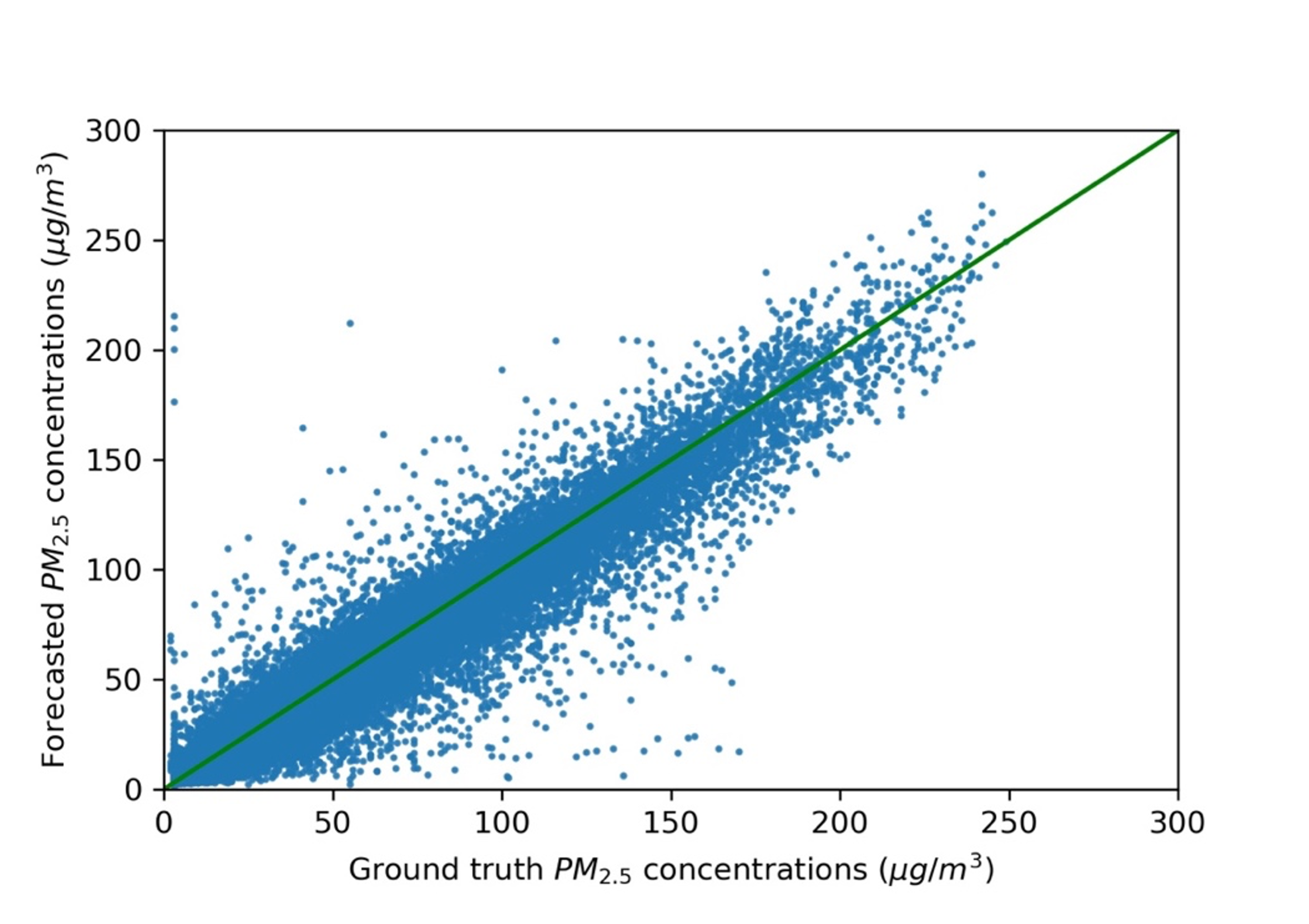}
  \caption{Comparison of the 1-Hour PM\textsubscript{2.5} Predicted Values by Deep-AIR and the Ground-truth PM\textsubscript{2.5} Values in (a) Hong Kong and (b) Beijing}
  \label{fig:forecast_scatter_plot} 
\end{figure*}

Although our proposed model has achieved the best performance for forecasting one-hour air quality, the error rate varied across different pollutants. Figure \ref{fig:forecast_scatter_plot} shows the relationship between the one-hour forecasts and the ground truths of PM\textsubscript{2.5} forecasted values in Hong Kong and Beijing. As shown on the two scatter plots, the points are distributed closely around the identity line. The R-squared value between the forecasted values and ground truths is 93\% in Hong Kong and 90\% in Beijing, suggesting that the one-hour PM\textsubscript{2.5} forecast values are highly consistent with the ground truths. In addition to PM\textsubscript{2.5} values, Table \ref{table:pollutant} shows the error rate of our proposed model when forecasting the one-hour pollution levels for each air pollutant. Results show that our proposed model performs best when forecasting PM\textsubscript{2.5} and PM\textsubscript{10} values in Hong Kong and CO values in Beijing. The one-hour forecasting error for PM\textsubscript{2.5} and PM10 is 15.8\% and 14.5\% in Hong Kong, respectively. The one-hour forecast error for CO values is 18.6\% in Beijing. However, in both cities, our proposed model fails to forecast O\textsubscript{3} values with very satisfactory accuracy, probably because the variation of O\textsubscript{3} values is not as regular as that of other pollutants. Given that sudden changes in O\textsubscript{3} values may occur from time to time, the proposed deep learning model may fail to learn the relevant patterns based on single historical data.

\begin{table}[t]
    \centering
    \caption{Error Rates (MAPEs) of 1-hour Air Pollution Predicated Values by Deep-AIR for Each Air Pollutant in Hong Kong and Beijing}
    \begin{tabular}{|l|l|l|l|l|l|l|}
    \hline
    \textbf{City / Pollutant} & PM\textsubscript{2.5} & PM\textsubscript{10} & NO\textsubscript{2} & SO\textsubscript{2} & O\textsubscript{3} & CO \\ \hline
    Hong Kong & 15.8 & \textbf{14.5} & 26.1 & 20.9 & 37.1 & N.A. \\ \hline
    Beijing & 24.5 & 25.4 & 24.7 & 23.3 & 31.8 & \textbf{18.6} \\ \hline
    \end{tabular}
    \label{table:pollutant}
\end{table}

\begin{figure*} 
  \centering
  [a] \includegraphics[width=0.3\linewidth]{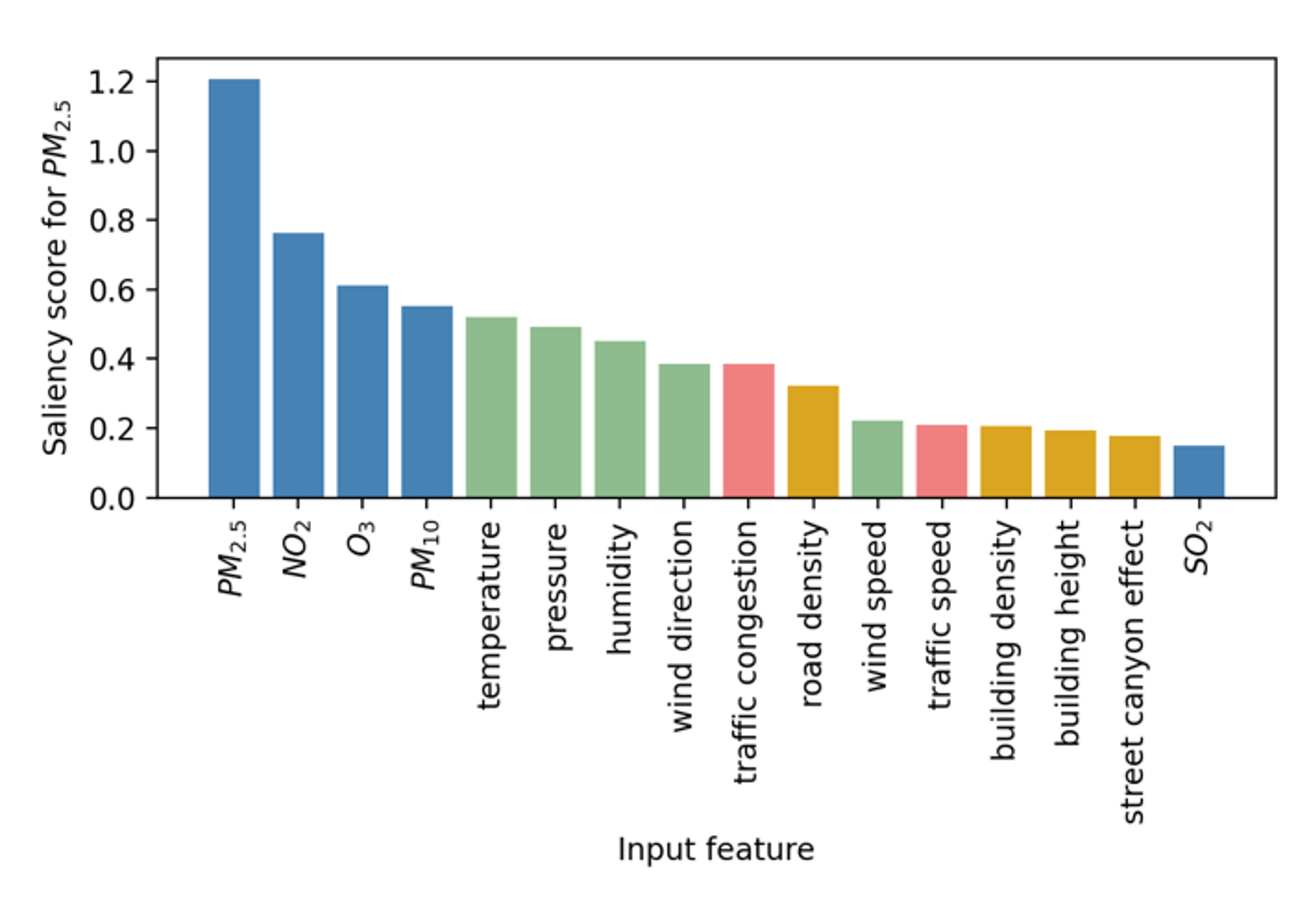}
  \hfill
  [b] \includegraphics[width=0.3\linewidth]{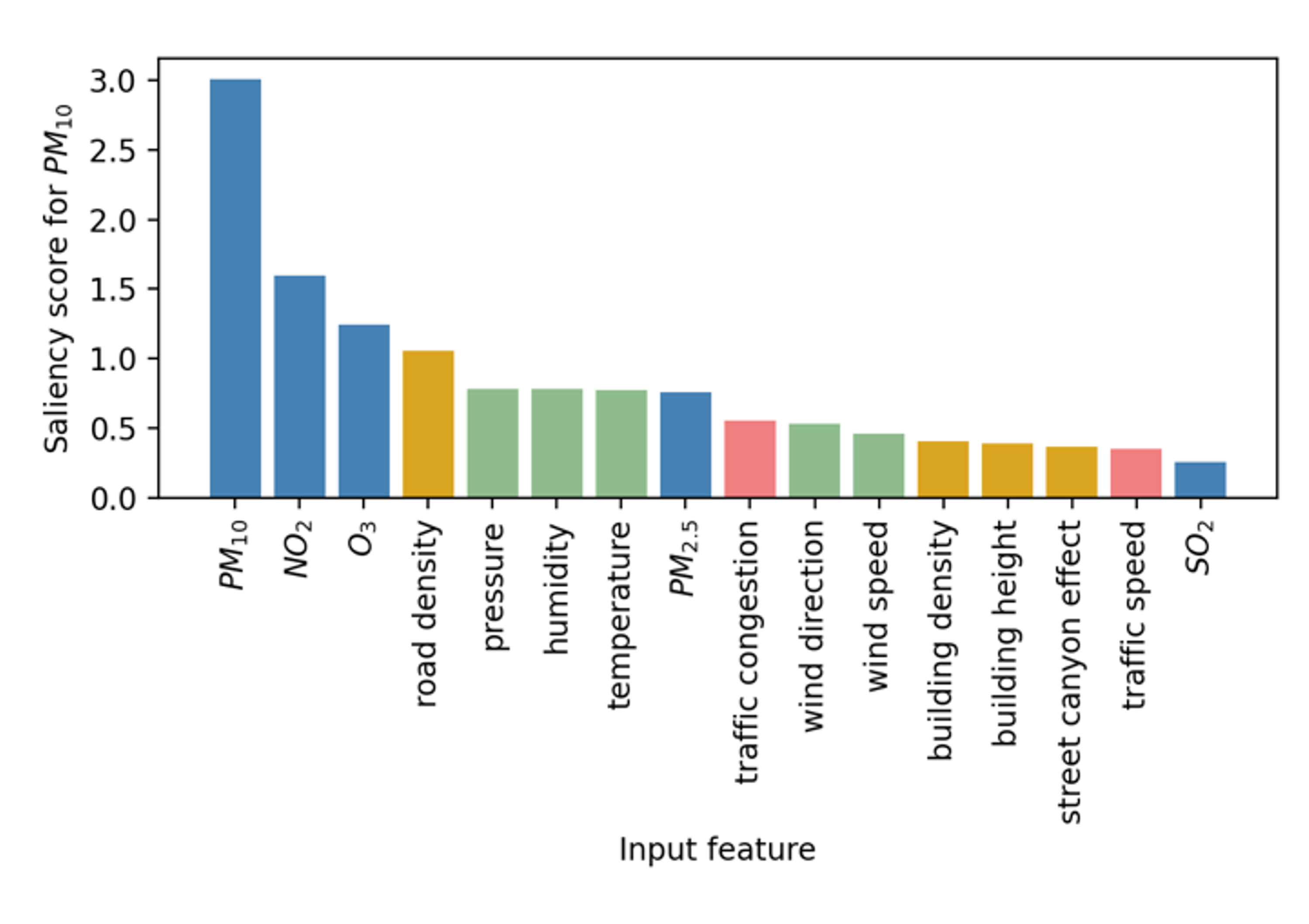}
  \hfill
  [c] \includegraphics[width=0.3\linewidth]{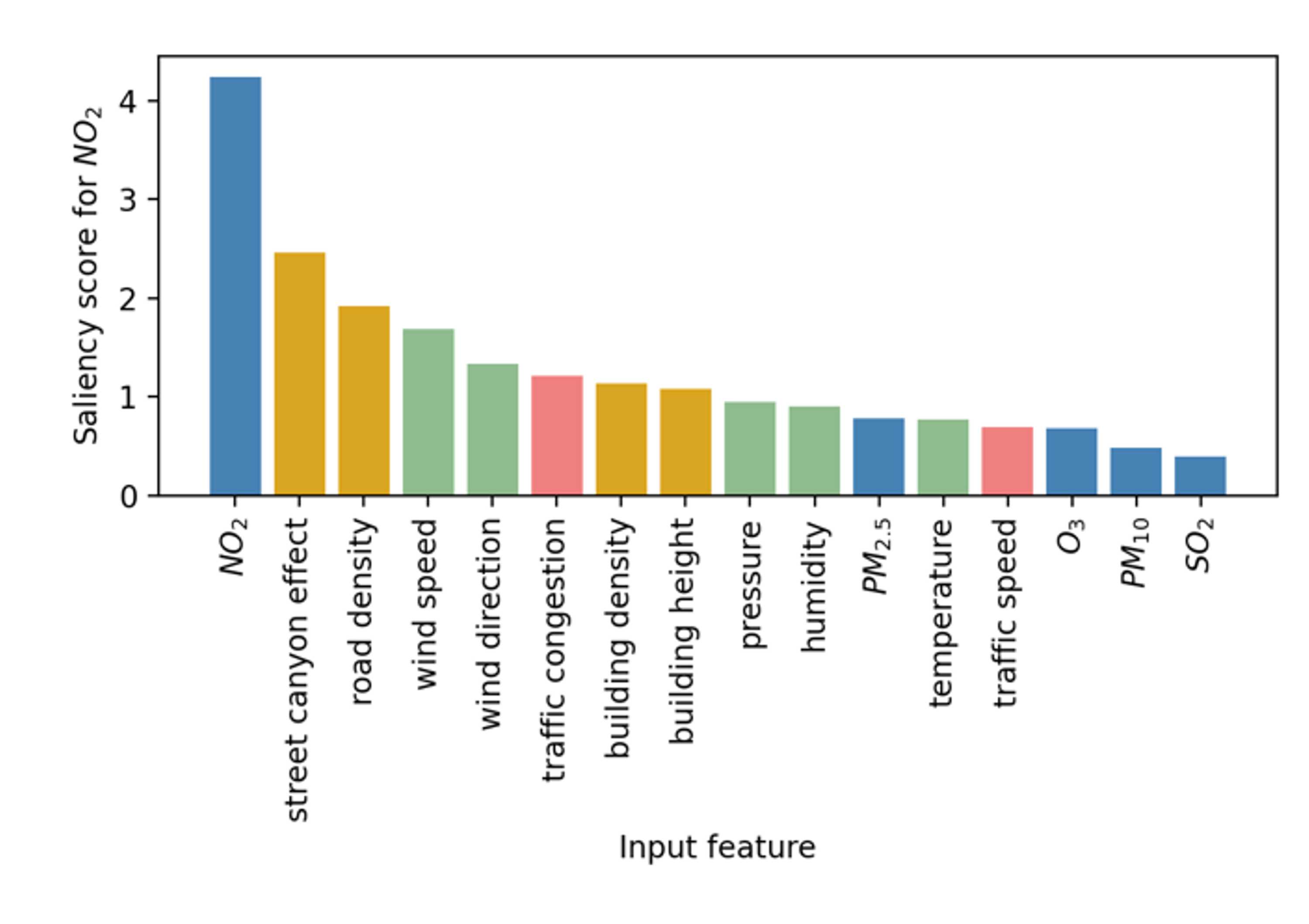}
  \hfill
  [d] \includegraphics[width=0.3\linewidth]{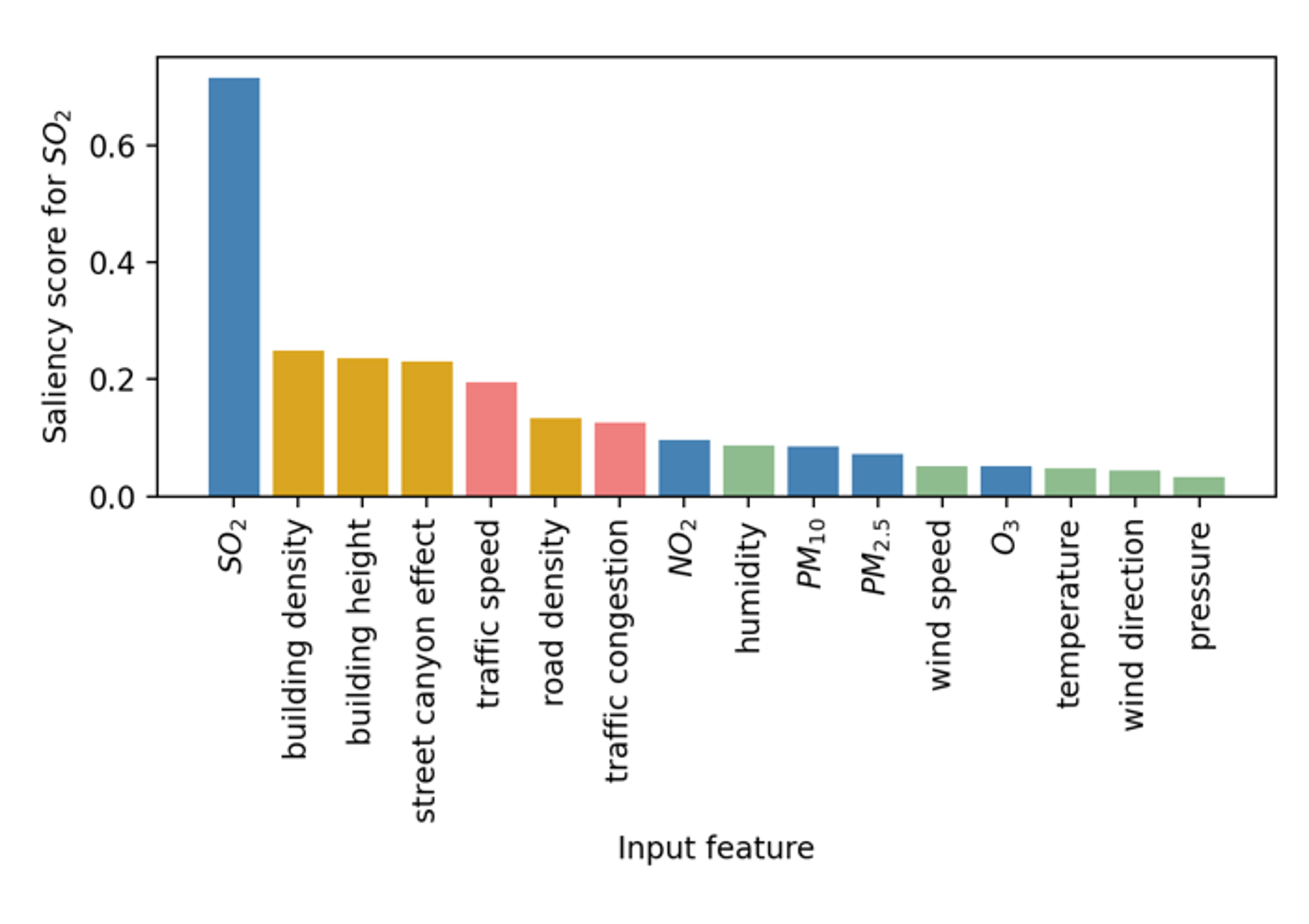}
  \hfill
  [e] \includegraphics[width=0.3\linewidth]{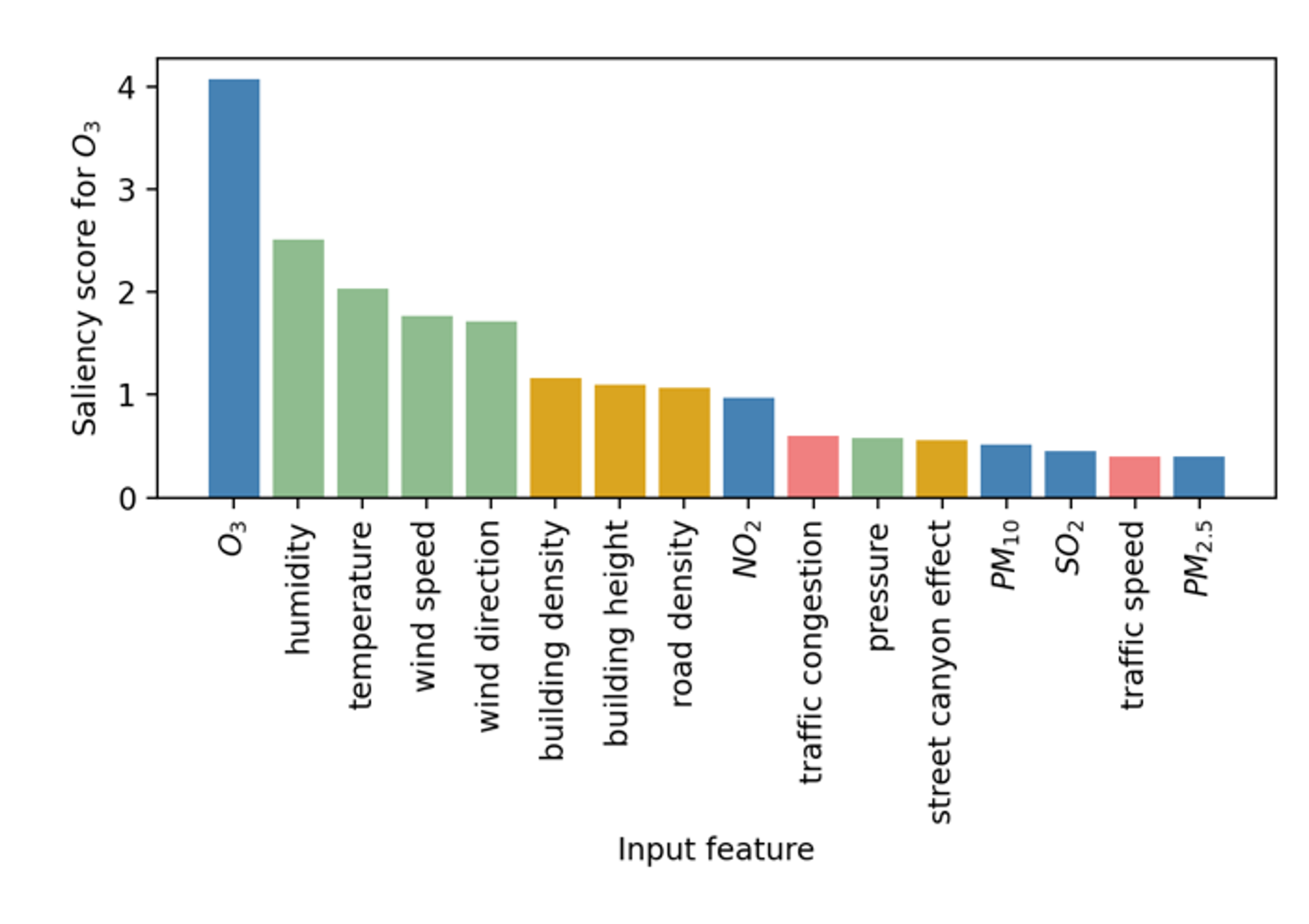}
  \hfill
  \phantom{[f]} \includegraphics[width=0.3\linewidth]{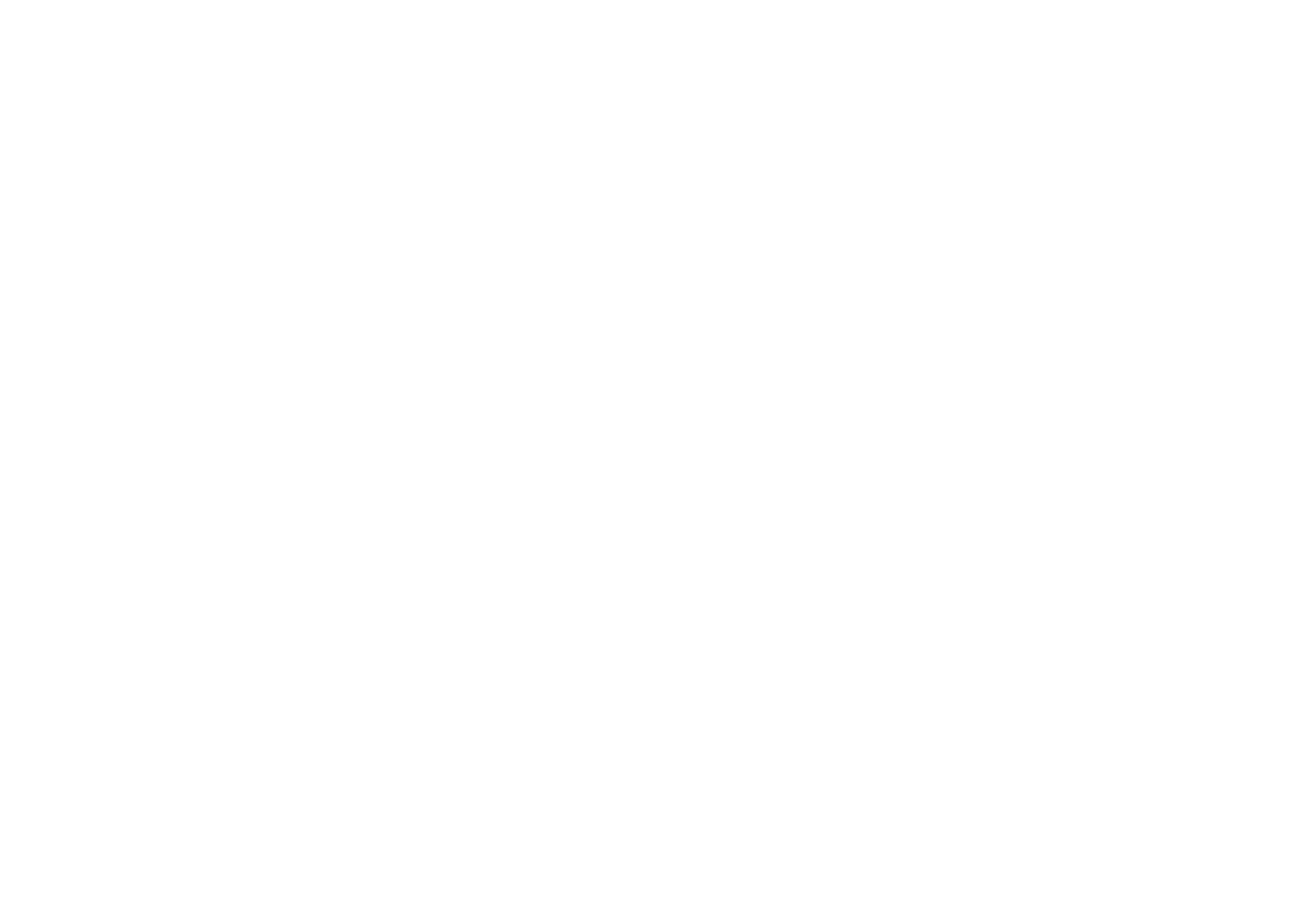}
  \caption{Saliency Scores of Different Input Features Contributing to Air Pollution Forecast in Hong Kong: (a) for PM\textsubscript{2.5}, (b) for PM\textsubscript{10}, (c) for NO\textsubscript{2}, (d) for SO\textsubscript{2}, and (e) for O\textsubscript{3}}
  \label{fig:saliency_hk} 
\end{figure*}

\begin{figure*} 
  \centering
  [a] \includegraphics[width=0.3\linewidth]{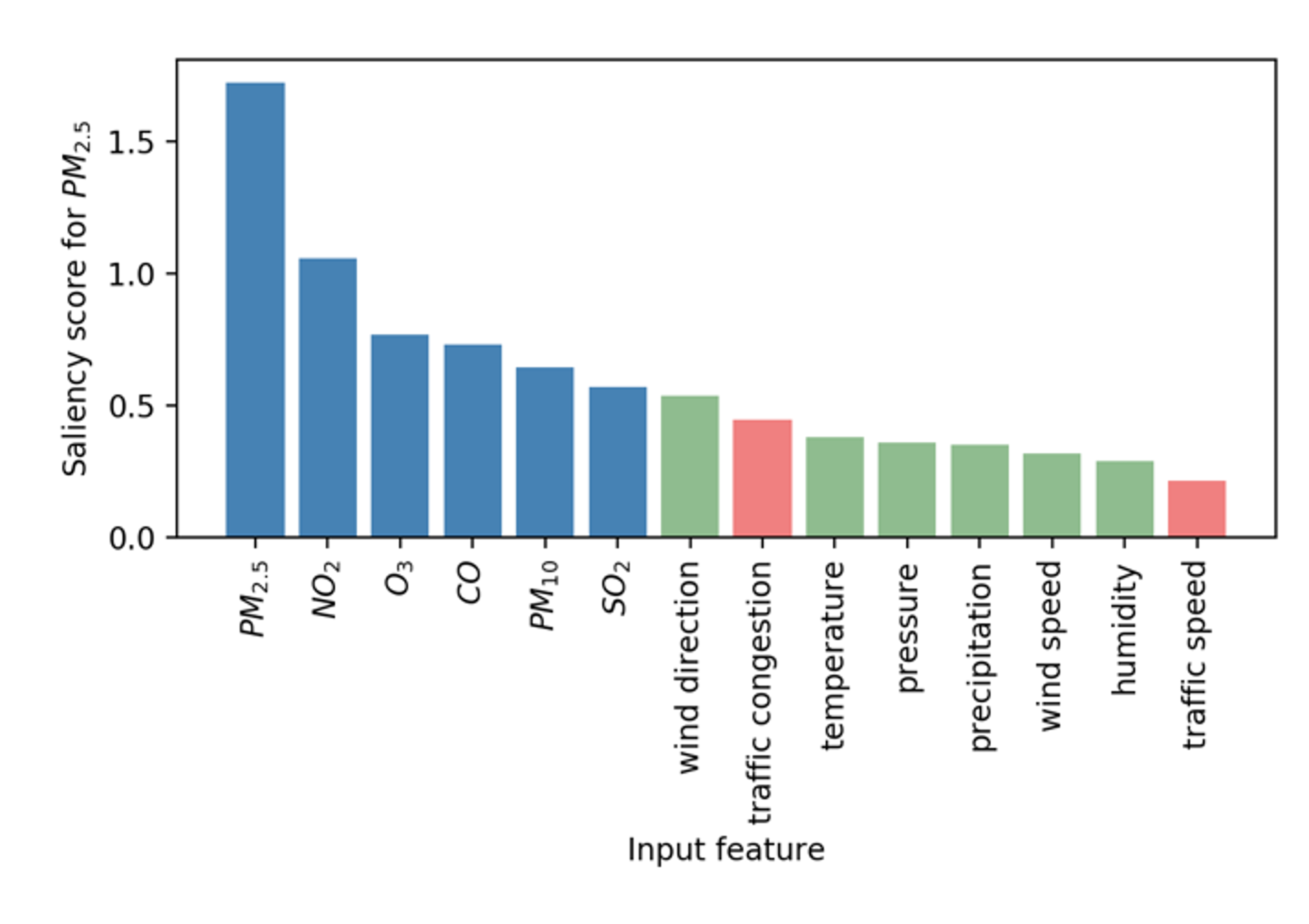}
  \hfill
  [b] \includegraphics[width=0.3\linewidth]{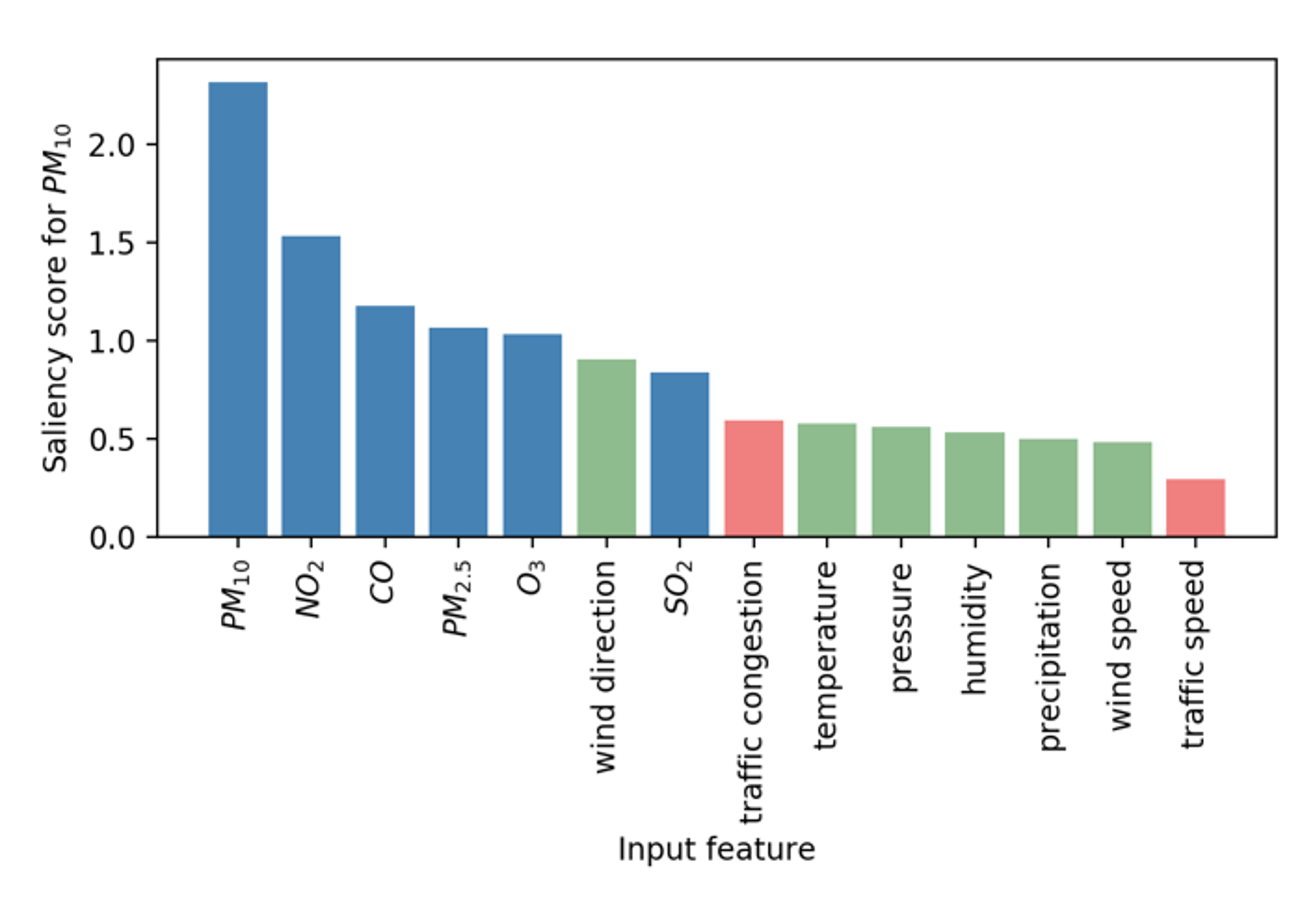}
  \hfill
  [c] \includegraphics[width=0.3\linewidth]{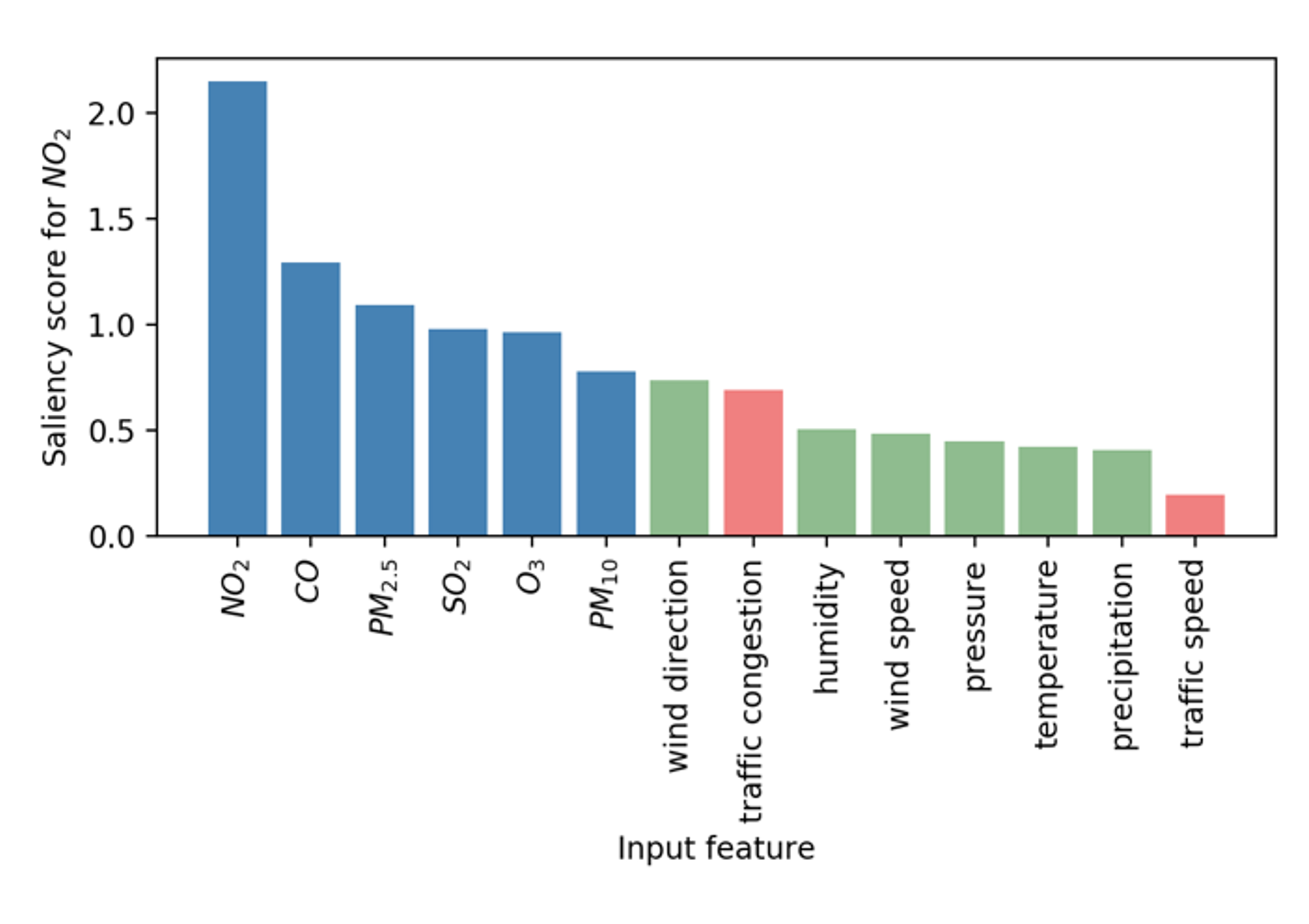}
  \hfill
  [d] \includegraphics[width=0.3\linewidth]{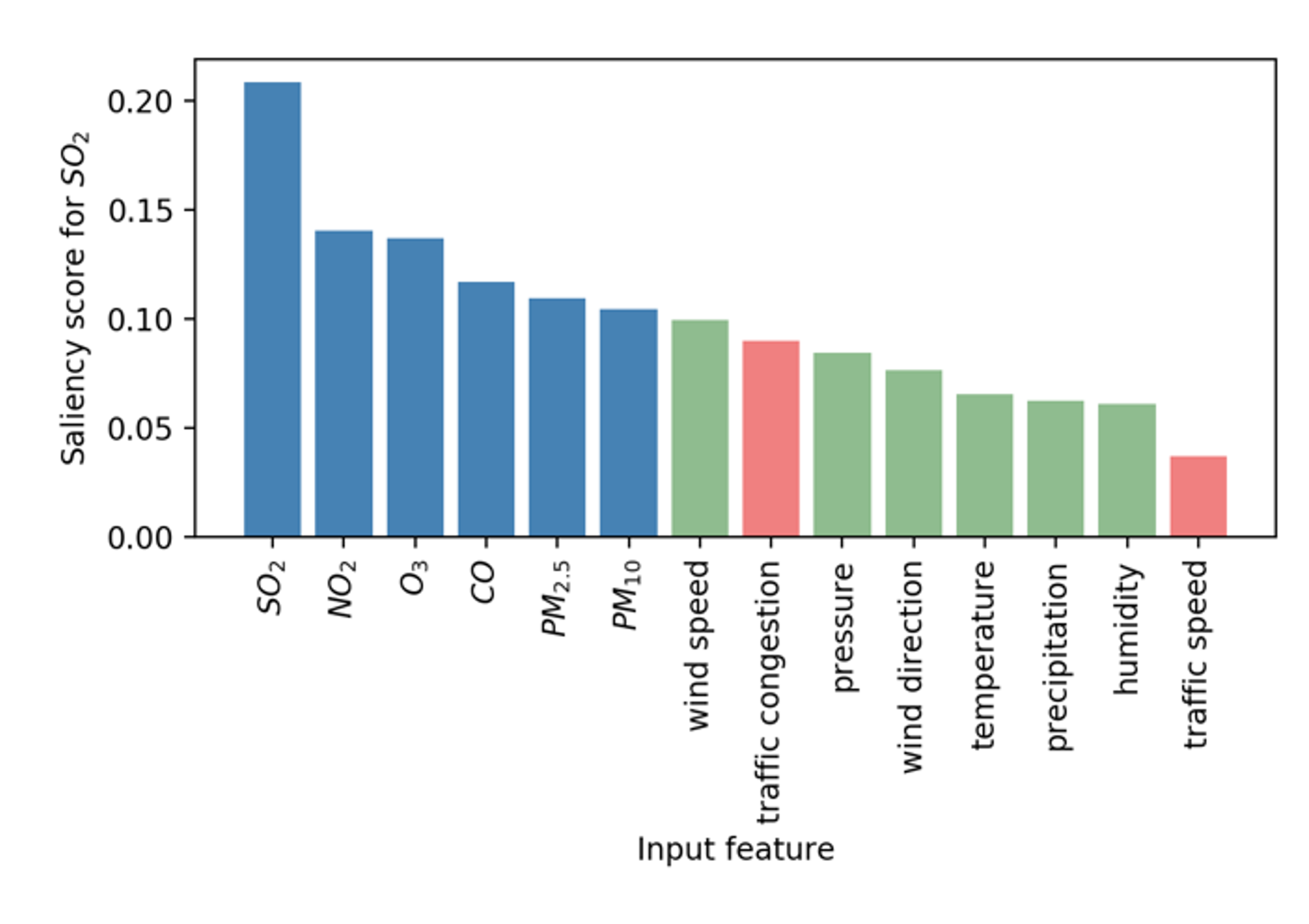}
  \hfill
  [e] \includegraphics[width=0.3\linewidth]{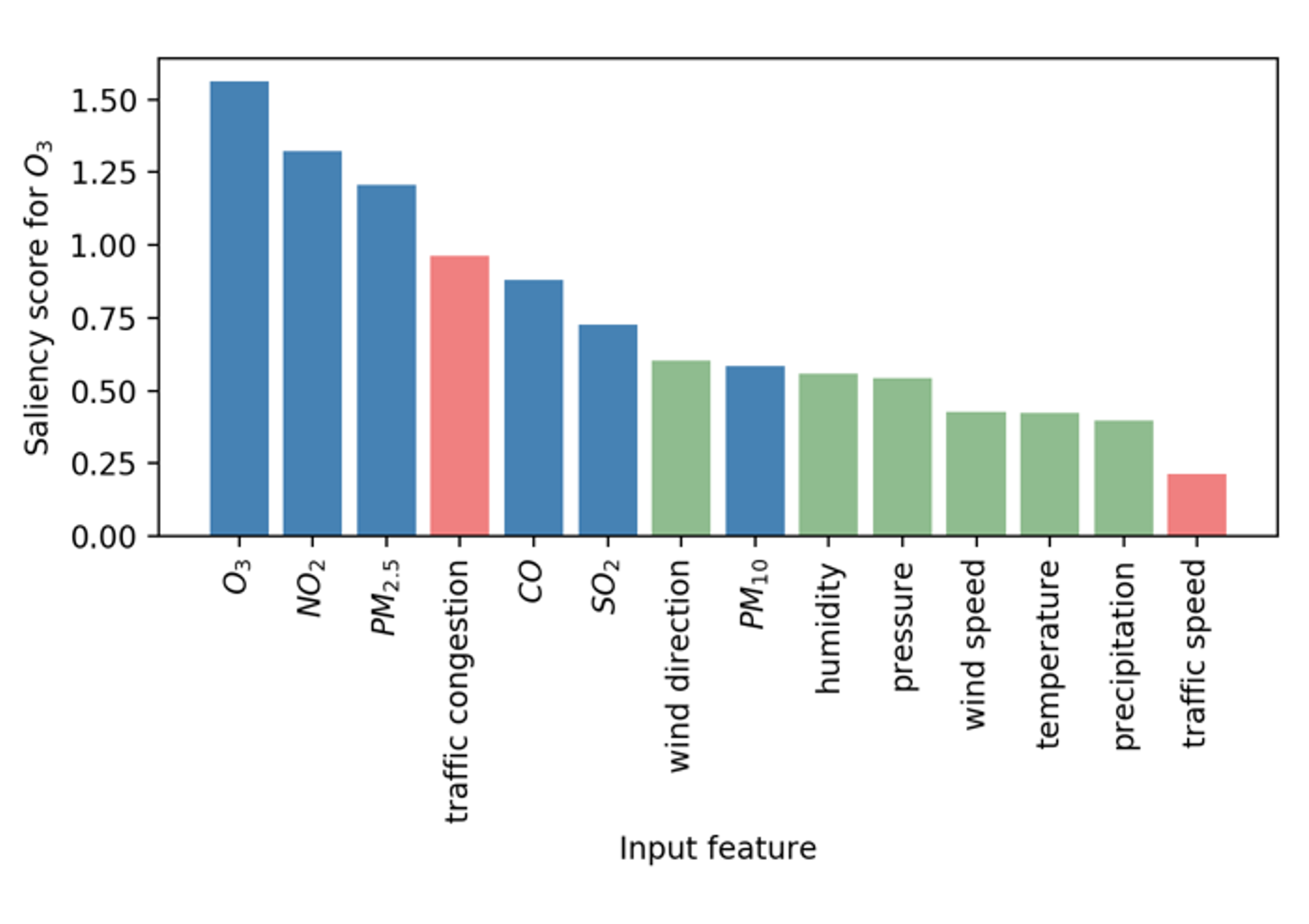}
  \hfill
  [f] \includegraphics[width=0.3\linewidth]{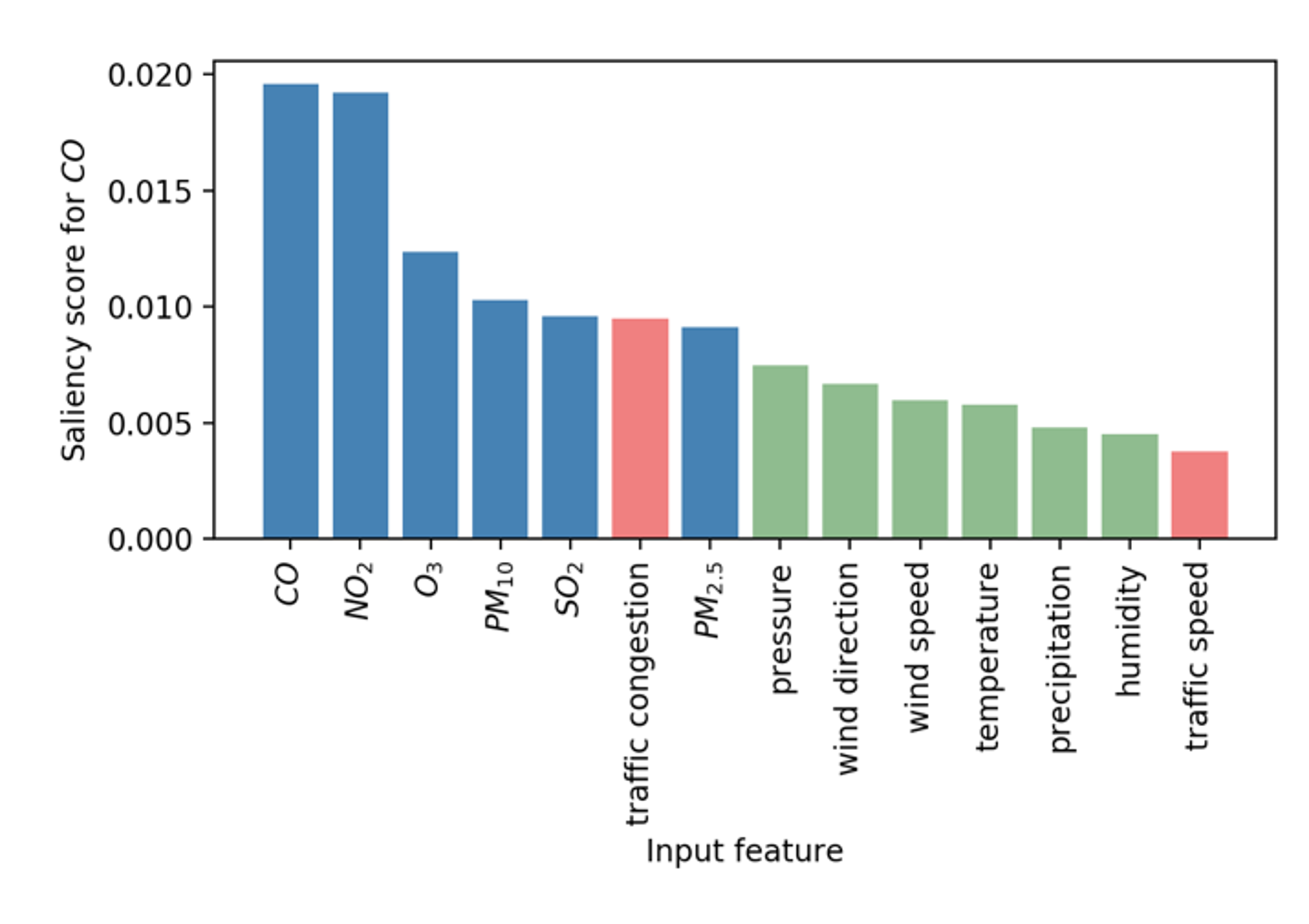}
  \caption{Saliency Scores of Different Input Features Contributing to Air Pollution Forecast in Beijing: (a) for PM\textsubscript{2.5}, (b) for PM\textsubscript{10}, (c) for NO\textsubscript{2}, (d) for SO\textsubscript{2}, (e) for O\textsubscript{3}, and (f) for CO}
  \label{fig:saliency_beijing} 
\end{figure*}

Furthermore, the saliency analysis has revealed more insights on which features are important for urban air pollution modeling. Figures \ref{fig:saliency_hk} and \ref{fig:saliency_beijing} illustrate the saliency scores of different input features contributing to air pollution forecasting in Hong Kong and Beijing. In general, the historical air pollution data are the most influential features, indicating that air pollution forecasts in the next hours are most relevant to the air pollution levels in recent hours. Among the air pollution data, each pollutant has always been the most important influential factor for forecasting itself, while NO\textsubscript{2} is often the most influential factor when forecasting non-NO\textsubscript{2} pollutants. Moreover, the importance of urban proxy data varied across pollutants and cities. On the one hand, the street canyon-related features play a more important role in predicting the air quality in Hong Kong. For predicting NO\textsubscript{2} and SO\textsubscript{2} values, urban morphological features, including the street canyon effect (binary indicator), road density, building density, and building height, are the most important proxy features, carry a higher impact than the meteorology and traffic features, and historical observations of other pollutants. For predicting PM\textsubscript{10} values, historical NO\textsubscript{2} and O\textsubscript{3} pollution levels and road density are the most salient factors. However, for PM\textsubscript{2.5} and O\textsubscript{3} forecasts, the street canyon-related features are less salient, and meteorological features are more important in predicting these pollutants. On the other hand, given that urban morphology features are absent from the Beijing dataset, traffic congestion and wind direction become the two most important urban proxy features in predicting Beijing's air quality. However, the meteorology and traffic features are less salient than the historical air pollution features in Beijing air pollution prediction.

\section{Discussions and Future Work}\label{sec:discussions}
This study aims to estimate fine-grained air pollution at the city-wide level in the current hour, while forecasting air pollution at monitoring stations in the short term (one hour ahead) and the long term (24 hours ahead). Using Hong Kong and Beijing as the case studies, our proposed novel deep learning framework, Deep-AIR, utilizes a large amount of readily available urban proxy data, while using 1x1 convolution layers to better capture cross-feature spatial interaction between air pollution and various important urban dynamic features, including meteorology, traffic conditions, and urban morphology (road density, building density/height, and street canyon).

The experimental results show that our proposed framework has achieved better performance than the baselines, including statistical and deep learning models, in fine-grained air pollution estimation and air pollution forecast (see Tables \ref{table:air_pollution_estimation}-\ref{table:air_pollution_forecast_beijing}). Compared to ARIMA and LSTM, other models that carry spatial structures, including ConvLSTM, ResNet-LSTM, and DeepAIR, have generally achieved a higher accuracy, suggesting the importance of taking into account the spatial information in air quality prediction. Our proposed model and ResNet-LSTM, each with separate components for modeling the spatio-temporal correlation, have performed better than the ConvLSTM, integrating the convolutional method into the LSTM unit directly. This is probably because ConvLSTM may fail to capture the long-term temporal dependence of high-dimensional urban dynamics directly. In contrast, the LSTM component of our proposed model or ResNet-LSTM takes the one-dimensional high-level representation (extracted from CNN) as the input. Our proposed model has also outperformed ResNet-LSTM without 1x1 convolution layers, suggesting that CNN with 1x1 convolution layers can learn better the spatial representation of the cross-urban-dynamic interaction.

Our ablation analysis has further confirmed that using 1x1 convolution layers can better capture the spatial interaction between different types of urban dynamics for forecasting air pollution (see Tables \ref{table:air_pollution_forecast_hk} and \ref{table:air_pollution_forecast_beijing}). On the one hand, compared to the air pollution forecast models using air pollution features only, the performance improvement of our proposed model is having the most extensive scale as compared to all baselines, when more features, including meteorology, traffic, and urban morphology have been added. This observation indicates that our proposed air pollution forecast model has the best potential to utilize more domain-specific urban proxy data. On the other hand, our proposed air pollution forecast model still outperforms the baseline models using fewer input data types. One exception being that when using air pollution data only, the prediction error of our proposed model was slightly higher than or equal to its counterpart without the 1x1 convolution layers (ResNet-LSTM), likely attributable to zero cross-urban-dynamic interaction. Although the image processing community has already utilized 1x1 convolution layers for modeling cross-channel correlations \cite{chollet2017xception}, few studies have adopted this structure in deep learning-based air quality modeling. In a recent study, a 1x1 convolution layer was applied to the final output of a ConvLSTM model for forecasting city-wide air pollution \cite{le2020spatiotemporal}. However, as shown in the results of \cite{le2020spatiotemporal}, ConvLSTM's prediction error increased when all input features, including meteorology and traffic conditions, were included, suggesting that unlike our proposed approach, the combination of ConvLSTM with the 1x1 convolution layer is less capable of capturing the spatial interaction between different types of urban dynamics.

Moreover, our experimental results show that estimating real-time fine-grained air pollution is more challenging than forecasting one-hour air pollution at monitoring stations. The lowest error rate of our proposed model for one-hour-ahead air pollution forecast was 22.8\% and 24.7\% in Hong Kong and Beijing, respectively. For fine-grained air pollution estimation in the current hour, the lowest error rate of our proposed model is 32.4\% and 35.0\% in Hong Kong and Beijing, respectively. Compared to 1-hour air pollution forecasts at monitoring stations where historical air quality information was available, the error rate is significantly higher when estimating real-time air pollution levels in locations without local air quality information. This suggests that the sparsity of ground truths has limited the accuracy of fine-grained air pollution estimation. Such finding is consistent with a previous data-driven air quality modeling study where the results of both fine-grained air pollution estimations and 1-hour air pollution forecasts were compared \cite{zhao2017incorporating}. Nevertheless, it remains challenging to provide accurate long-term air pollution forecasts, even though historical air quality measurements are available. The lowest error rate of our proposed model for 24-hour-ahead air pollution forecast is 33.9\% and 36.5\% in Hong Kong and Beijing, respectively. Forecasting long-term air pollution is inaccurate due to various issues such as the accumulation of errors during multi-step forecasting \cite{zhou2019explore} and the lack of future information (proxy data) such as weather and traffic forecast \cite{yi2020predicting}.

Furthermore, our saliency analysis has highlighted the importance of domain-specific features for urban air quality modeling (see Figures \ref{fig:saliency_hk} and \ref{fig:saliency_beijing}). Such findings can improve the interpretability of our proposed framework, while providing more insights for deep learning-based air quality modeling to account for the most important domain-specific features. The saliency scores obtained from our proposed model are consistent with domain knowledge. On the one hand, the urban morphology features, including street canyon effect (binary indicator), road density, building density, and building height, are important factors in predicting NO\textsubscript{2} pollution in Hong Kong, a city characterized by closely packed high-rise buildings and heavy road traffic. This finding is consistent with previous studies that adopted land-use regression to model the spatial variability of NO\textsubscript{2} values in urban areas \cite{tang2013using}. Unlike NO\textsubscript{2} values that are sensitive to traffic conditions, the urban morphological features are less salient in predicting PM or O\textsubscript{3} values in Hong Kong, probably because a large proportion of the PM pollution within the city is associated with the long-range PM transportation from other regions outside of the city \cite{tang2013using}, whereas O\textsubscript{3} emissions are generated from secondary sources, dependent not only on traffic emissions but also meteorological conditions \cite{vardoulakis2003modelling}. Further, the saliency scores of urban morphological features for predicting SO\textsubscript{2} values are small, probably due to the fact that the most important sources of SO\textsubscript{2} emissions in Hong Kong, including electricity generation and marine transportation, are not taken into account in the urban proxy data. Nevertheless, real-time traffic features are less salient for all air pollutants in Hong Kong, likely attributable to the lack of traffic data with high coverage. Further investigation on the importance of traffic conditions is needed after large-scale traffic data are collected in Hong Kong. On the other hand, traffic congestion and wind direction are the two most important features in predicting air pollution in Beijing, a city facing significant air pollution problems due to local vehicular emissions and regional PM\textsubscript{2.5} transportation from nearby cities. Although the urban morphological data are unavailable in the Beijing dataset, previous physical modeling studies demonstrated the importance of canyon geometries in estimating high-resolution NO\textsubscript{2} pollution in Beijing \cite{fu2017effects}. Given that the urban morphological features have already played an important role in predicting traffic-related pollution in Hong Kong, relevant data can be collected in Beijing to improve the performance of deep learning-based air quality prediction in the city.

Several challenges remain to be overcome in the current study. First, the missing/sparse urban proxy data have limited the performance of our proposed model. The geographical coverage of the collected traffic data in Hong Kong remains low, whereas urban morphology data are unavailable in Beijing. In the future, we will collect real-time travel time data in Hong Kong from Google Maps to complement the traffic data collected from the official source \cite{he2019comparative}. We will also collect street-view images in Beijing from Baidu Maps to estimate urban canyon geometries using CNN models \cite{hu2020classification}. Second, the importance of different domain-specific features has yet to be considered in our proposed model. In the future, we will investigate more advanced deep learning model structures, such as attention neural networks and graph-based neural networks, to tailor-make model structures that utilize the most relevant features for air quality prediction. Third, the generalizability of our proposed fine-grained air pollution estimation model should be further investigated. In our experiments, the dataset split for evaluating the fine-grained air pollution estimations was based a random split. However, this may lead to low model generalization ability when testing in new locations, given that all monitoring stations were included during model training. In the future, the evaluation of fine-grained air pollution estimation can be based on a hold-out air quality station, which is removed during model training and is used for model evaluation only. Given that the number of air quality monitoring stations remains small in Hong Kong and Beijing, we will adopt testing methods that can better use the collected ground truths, such as leave-one-out cross-validation, to further evaluate the performance of our proposed model for city-wide fine-grained air pollution estimation.

Finally, based on our proposed hybrid deep learning framework, forecasting fine-grained air pollution at the city-wide level would be a good research direction. As compared to fine-grained air pollution estimation in the existing hour, fine-grained air pollution forecast is more complicated because air quality measurements over both space and time are sparse. Several deep learning studies have attempted to forecast the spatio-temporal variation of air pollution across different parts of the city. Fan et al. \cite{fan2017spatiotemporal}  constructed a fine-grained PM\textsubscript{2.5} forecast map, using an IDW-based spatial interpolation method, based on station-level forecasts predicted by an LSTM model. Along this line, Ma et al. \cite{ma2019spatiotemporal} constructed a fine-grained PM\textsubscript{2.5} forecast map by interpolating station-level forecasts from a bidirectional LSTM model, using an IDW-based layer with optimized parameters. Moreover, Qi et al. \cite{qi2018deep} proposed a deep feedforward neural network to estimate fine-grained air pollution in the current hour and future hours, utilizing a feature selection layer and the spatial and temporal information obtained from neighboring air quality and meteorology observations. Nevertheless, given that these studies have not adopted advanced deep learning techniques such as CNN models that can better account for the spatial structure of data, the complex spatial dependence of air pollution and urban proxy data is yet to be fully addressed. More recently, Le et al. \cite{le2020spatiotemporal} proposed a ConvLSTM model to forecast fine-grained air pollution at the city-wide level, but they did not consider urban morphology features. Until now, how to utilize advanced spatio-temporal models to better account for important spatial features and their interactions in forecasting fine-grained air pollution throughout a city remains to be investigated further. The air pollution forecast model in our current framework has been trained and tested at locations where historical ground-truth air quality measurements are available. In the future, we will extend our proposed air pollution forecast model to provide fine-grained air pollution forecasts in areas not covered by air quality monitoring stations.

\section{Conclusions}\label{sec:conclusions}
This study proposes a hybrid deep learning framework (Deep-AIR) for fine-grained air quality estimation at the city-wide metropolitan level in the current hour and air pollution forecast at monitoring stations up to 24 hours ahead. To the best of our knowledge, this study is the first of its kind to introduce domain-specific features related to the street canyon effect, including building density and building height, for deep learning-based air quality model estimation and prediction. To better exploit the important urban dynamic features and their spatial interaction for predicting air pollution in high-density metropolitan cities, our proposed framework utilizes 1x1 convolutional layers in CNN to account for the complex spatial interaction between air pollution and other domain-specific urban dynamic features. Experimental results show that our proposed Deep-AIR has outperformed the baselines in fine-grained air pollution estimation and air pollution forecast. In Hong Kong, our proposed model has achieved 67.6\%, 77.2\%, and 66.1\% accuracy for fine-grained air pollution estimation, 1-hour air pollution forecast, and 24-hour air pollution forecast, respectively. Similarly, our proposed model has achieved 65.0\%, 75.3\%, and 63.5\% accuracy for the same three categories in Beijing. In addition, our saliency analysis has shown that for Hong Kong, urban morphology, such as street canyon and road density, are the best estimators for NO\textsubscript{2}, while meteorology is the best estimator for PM\textsubscript{2.5}. Such findings can shed new lights on deep learning-based air quality modeling studies, given that the street canyon effect has largely been overlooked. In the future, we will collect more domain-specific data at a large scale, including traffic conditions data in Hong Kong and urban morphology data in Beijing. We will improve the accuracy of our proposed framework by tailor-making model structures that account for high-saliency domain-specific features for air pollution prediction. We will also extend our proposed framework to provide city-wide fine-grained air pollution forecasts and conduct a more comprehensive test to evaluate the generalizability of city-wide fine-grained air pollution prediction.

\ifCLASSOPTIONcompsoc
  \section*{Acknowledgments}
\else
  \section*{Acknowledgment}
\fi

This research is supported in part by the Theme-based Research Scheme of the Research Grants Council of Hong Kong, under Grant No. T41-709/17-N. We also thank Microsoft for providing cloud computing services under Microsoft Azure. We would like to acknowledge the Environmental Protection Department of the HKSAR Government (HKEPD), the Hong Kong Observatory (HKO), and the Transport Department of the HKSAR Government (HKTD), for publicizing air quality, meteorology, and traffic data of Hong Kong, respectively. We also acknowledge the Lands Department of the HKSAR Government (HKLD) for providing the information on roads and buildings in Hong Kong. We would like to acknowledge the Beijing Municipal Environment Monitoring Center and the National Meteorological Information Center, China, for publicizing air quality and meteorological data of Beijing, respectively. We also acknowledge the map service of AutoNavi (Gaode) for providing real-time traffic data in Beijing.

\ifCLASSOPTIONcaptionsoff
  \newpage
\fi



%
\bibliographystyle{IEEEtran}
\bibliography{references}

%

\begin{IEEEbiography}[{\includegraphics[width=1in,height=1.25in,clip,keepaspectratio]{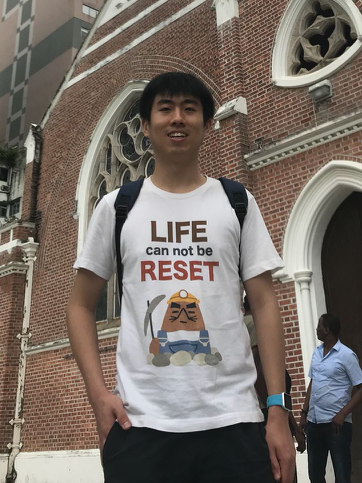}}]{Yang Han}
received the MSc degree in Computer Science (with distinction) from the University of Hong Kong (HKU) and the MPhil degree in Technology Policy from the University of Cambridge. Currently he is working towards PhD in AI and machine learning with applications on air pollution and health management at the Department of Electrical \& Electronic Engineering, HKU. His recent research specializes in spatio-temporal analysis and applications in environmental science and policy studies in China. He has published widely in Environmental Science and Policy, Data and Policy, and several IEEE journals.
\end{IEEEbiography}

\vskip 0pt plus -1fil

\begin{IEEEbiography}[{\includegraphics[width=1in,height=1.25in,clip,keepaspectratio]{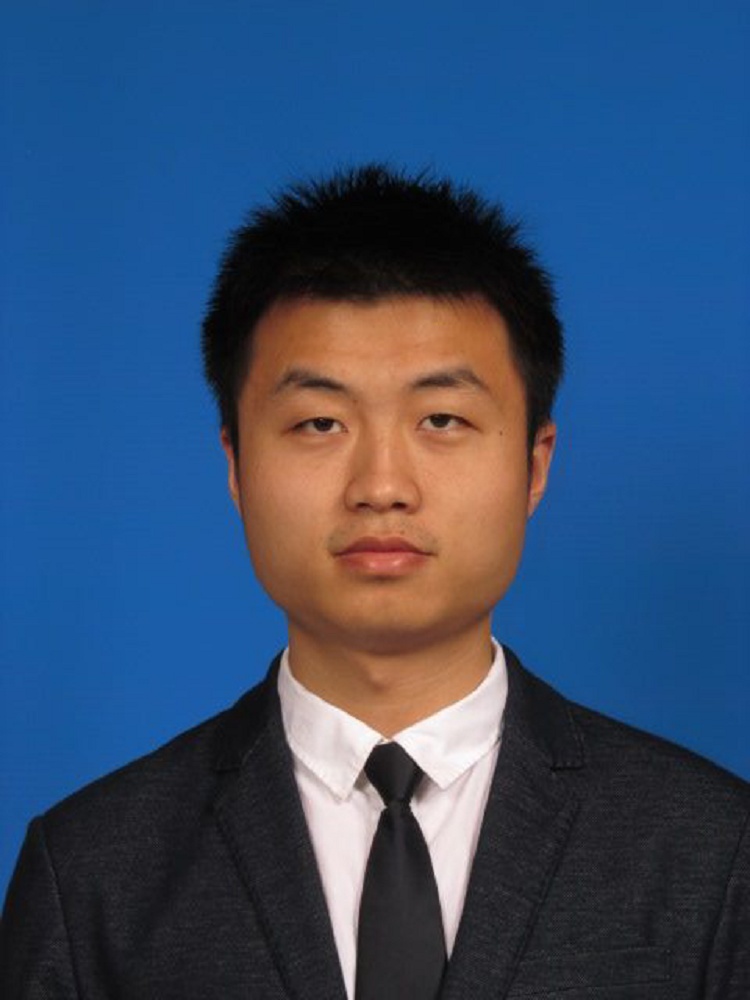}}]{Qi Zhang}
received the BE degree in electronic engineering and BEc degree in economics from Tsinghua University, Beijing, China, in 2017. He is working towards the PhD degree in the Department of Electrical \& Electronic Engineering, the University of Hong Kong. He is a holder of the Hong Kong PhD Fellowship. His research interests include deep-learning and its applications on air pollution, spatio-temporal data analysis, and urban computing.
\end{IEEEbiography}

\begin{IEEEbiography}[{\includegraphics[width=1in,height=1.25in,clip,keepaspectratio]{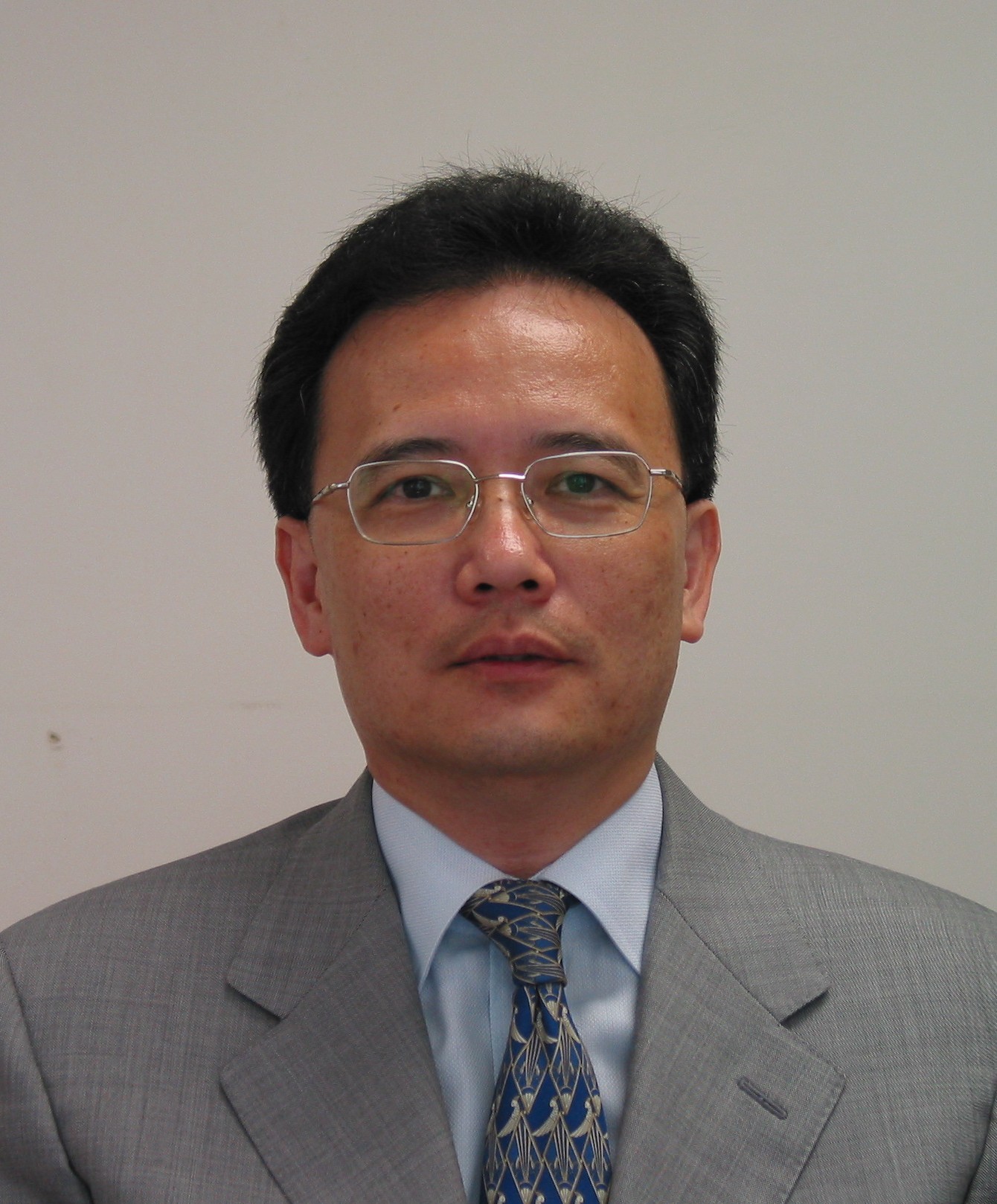}}]{Victor O.K. Li}
received SB, SM, EE and ScD degrees in Electrical Engineering and Computer Science from MIT. Prof. Li is Chair of Information Engineering and Cheng Yu-Tung Professor in Sustainable Development at the Department of Electrical \& Electronic Engineering (EEE) at the University of Hong Kong. He is the Director of the HKU-Cambridge Clean Energy and Environment Research Platform, and of the HKU-Cambridge AI to Advance Well-being and Society Research Platform, which are interdisciplinary collaborations with Cambridge University. He was Visiting Professor in the Department of Computer Science and Technology at the University of Cambridge from April to August 2019. He was the Head of EEE, Assoc. Dean (Research) of Engineering and Managing Director of Versitech Ltd. He serves on the board of Sunevision Holdings Ltd., listed on the Hong Kong Stock Exchange and co-founded Fano Labs Ltd., an artificial intelligence (AI) company with his PhD student. Previously, he was Professor of Electrical Engineering at the University of Southern California (USC), Los Angeles, California, USA, and Director of the USC Communication Sciences Institute. His research interests include big data, AI, optimization techniques, and interdisciplinary clean energy and environment studies. In Jan 2018, he was awarded a USD 6.3M RGC Theme-based Research Project to develop deep learning techniques for personalized and smart air pollution monitoring and health management. Sought by government, industry, and academic organizations, he has lectured and consulted extensively internationally. He has received numerous awards, including the PRC Ministry of Education Changjiang Chair Professorship at Tsinghua University, the UK Royal Academy of Engineering Senior Visiting Fellowship in Communications, the Croucher Foundation Senior Research Fellowship, and the Order of the Bronze Bauhinia Star, Government of the HKSAR. He is a Fellow of the Hong Kong Academy of Engineering Sciences, the IEEE, the IAE, and the HKIE.
\end{IEEEbiography}

\vskip 0pt plus -1fil

\begin{IEEEbiography}[{\includegraphics[width=1in,height=1.25in,clip,keepaspectratio]{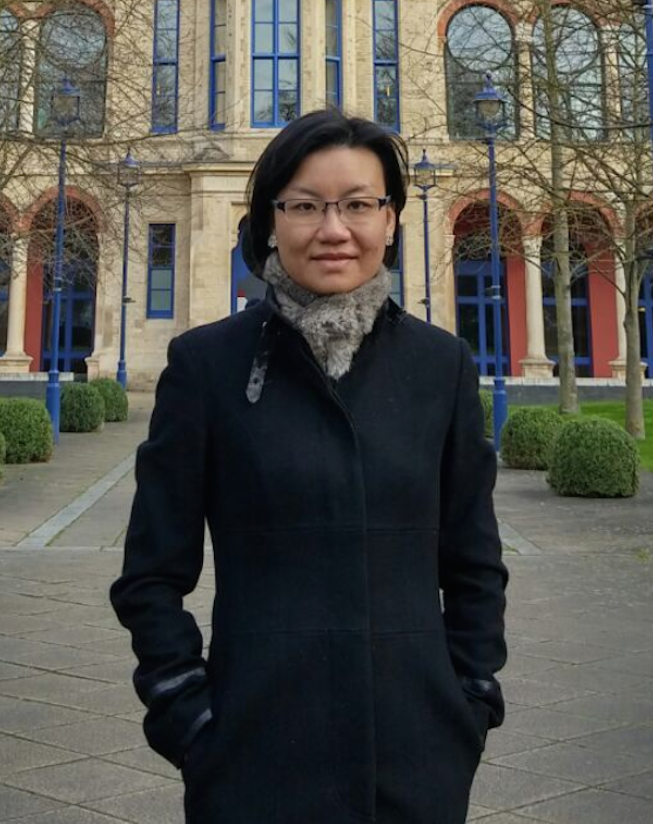}}]{Jacqueline C.K. Lam}
is Associate Professor at the Department of Electrical and Electronic Engineering, the University of Hong Kong. Since 2018 Jacqueline has co-established the HKU-AI WiSe with Prof. Victor OK Li, Chair of Information Engineering at the University of Hong Kong. She is the Co-Director of the HKU-Cambridge Clean Energy and Environment Research Platform, and of the HKU-Cambridge AI to Advance Well-being and Society Research Platform. She was the Hughes Hall Visiting Fellow, before she takes up the Visiting Senior Research Fellow and Associate Researcher in Energy Policy Research Group, Judge Business School, the University of Cambridge. Her research studies clean energy and environment using interdisciplinary approaches, with a special focus on China and the UK. Her recent research focuses on the use of big data and machine learning techniques to study personalized air pollution monitoring and health management. Her work is published in IEEE, Environment International, Applied Energy, Environmental Science and Policy, and Energy Policy. Jacqueline has received three times the research grants awarded by the Research Grants Council, HKSAR Government, during 2011-2017. The funded amount was USD 7.8M in PI/Co-PI capacity. Her recent research study, in joint collaboration with Yang Han and Victor OK Li on PM\textsubscript{2.5} pollution and environmental inequality in Hong Kong, has been published in Environmental Science and Policy, and widely covered by more than 30 local and overseas newspapers and TVs. She has recently co-organized the HKU-Cambridge AI for Social Good Symposium with Prof. Jonathan Crowcroft, FRS (CST Dept., Cambridge) and Prof. Victor OK Li (Clare Hall Life Fellow, Cambridge). Jacqueline also serves as the co-editor of a special issue on AI for environmental decision-making, published by Environmental Science and Technology, an Elsevier SCI journal.
\end{IEEEbiography}




\end{document}